\documentclass[10pt,twocolumn,letterpaper]{article}

\usepackage{iccv}
\usepackage{times}
\usepackage{epsfig}
\usepackage{graphicx}
\usepackage{amsmath}
\usepackage{amssymb}
\usepackage{caption}
\usepackage{subcaption}
\usepackage{booktabs} 
\usepackage{microtype}
\usepackage[symbol]{footmisc}
\usepackage{footnote}
\usepackage{algpseudocode}
\usepackage[flushleft]{threeparttable}

\DeclareMathOperator*{\argmax}{arg\,max}

\DeclareSymbolFont{bbold}{U}{bbold}{m}{n}
\DeclareSymbolFontAlphabet{\mathbbold}{bbold}

\usepackage[pagebackref=true,breaklinks=true,letterpaper=true,colorlinks,bookmarks=false]{hyperref}
\iccvfinalcopy 
\def\iccvPaperID{5296} 

\ificcvfinal\fi
\begin{document}

\title{Enhancing Adversarial Example Transferability with an Intermediate Level Attack}

\makeatletter
\newcommand{\printfnsymbol}[1]{%
  \textsuperscript{\@fnsymbol{#1}}%
}
\makeatother

\author{Qian Huang\thanks{Equal contribution.}\\
Cornell University\\
{\tt\small qh53@cornell.edu}
\and
Isay Katsman\printfnsymbol{1}\\
Cornell University\\
{\tt\small isk22@cornell.edu}
\and
Horace He\printfnsymbol{1}\\
Cornell University\\
{\tt\small hh498@cornell.edu}
\and
Zeqi Gu\printfnsymbol{1}\\
Cornell University\\
{\tt\small zg45@cornell.edu}
\and
Serge Belongie\\
Cornell University\\
{\tt\small sjb344@cornell.edu}
\and
Ser-Nam Lim\\
Facebook AI\\
{\tt\small sernam@gmail.com}
}

\maketitle

\begin{abstract}
   Neural networks are vulnerable to adversarial examples, malicious inputs crafted to fool trained models. Adversarial examples often exhibit black-box transfer, meaning that adversarial examples for one model can fool another model. However, adversarial examples are typically overfit to exploit the particular architecture and feature representation of a source model, resulting in sub-optimal black-box transfer attacks to other target models. We introduce the \textit{Intermediate Level Attack} (ILA), which attempts to fine-tune an existing adversarial example for greater black-box transferability by increasing its perturbation on a pre-specified layer of the source model, improving upon state-of-the-art methods. We show that we can select a layer of the source model to perturb without any knowledge of the target models while achieving high transferability. Additionally, we provide some explanatory insights regarding our method and the effect of optimizing for adversarial examples using intermediate feature maps. Our code is available at \url{https://github.com/CUVL/Intermediate-Level-Attack}.
\end{abstract}

\section{Introduction}
\label{intro}

\begin{figure}[htb]
    \centering
    \includegraphics[width=0.50\textwidth]{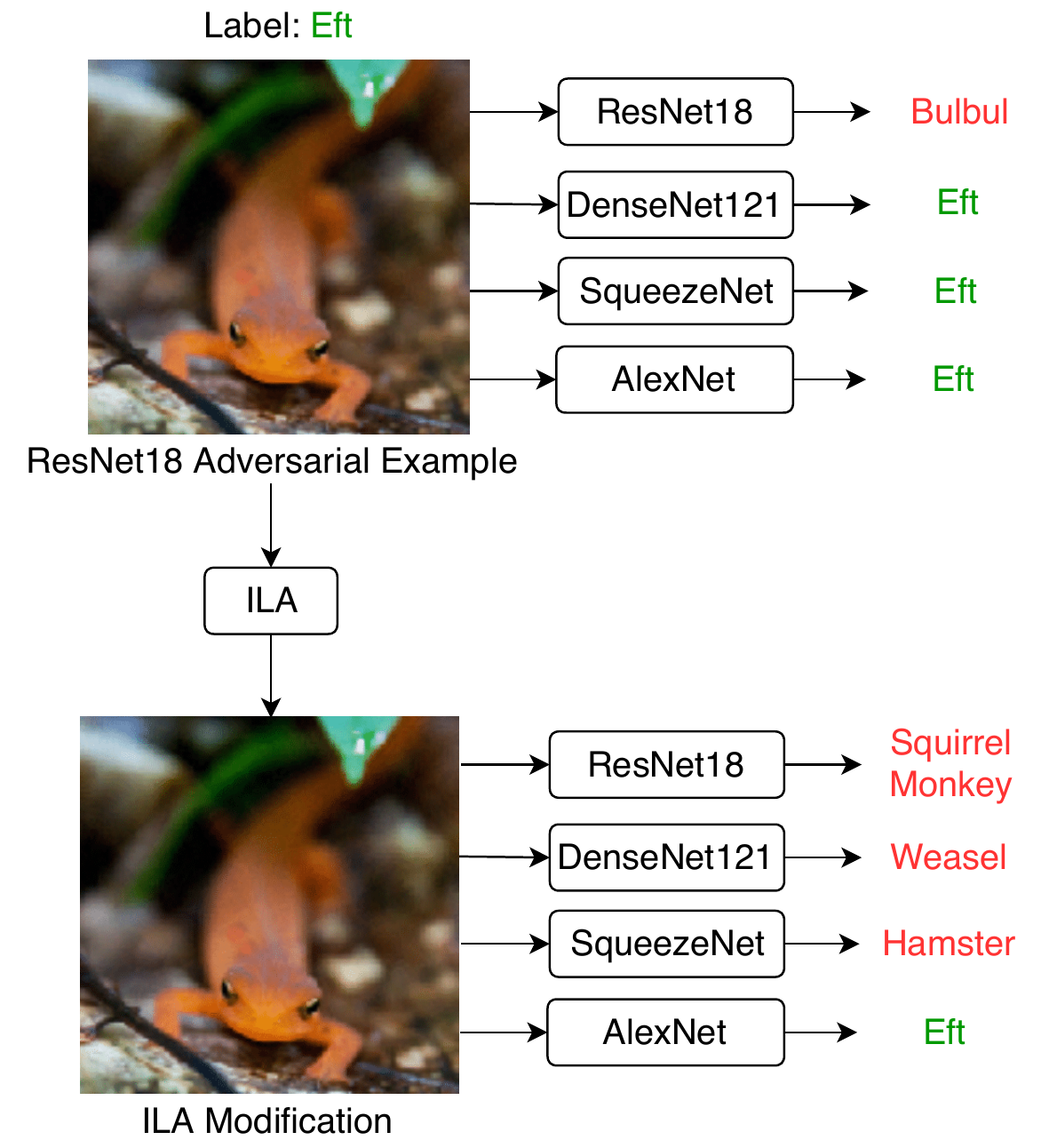}
    \caption{\label{fig:ilaintrofigure} An example of an ILA modification of a pre-existing adversarial example for ResNet18. ILA modifies the adversarial example to increase its transferability. Note that although the original ResNet18 adversarial example managed to fool ResNet18, it does not manage to fool the other networks. The ILA modification of the adversarial example is, however, more transferable and is able to fool more of the other networks.}
\end{figure}

Adversarial examples are small, imperceptible perturbations of images carefully crafted to fool trained models \cite{Szegedy2013IntriguingPO, Goodfellow2014ExplainingAH}. Studies such as \cite{Krizhevsky2012ImageNetCW} have shown that Convolutional Neural Networks (CNNs) are particularly vulnerable to such adversarial attacks. The existence of these adversarial attacks suggests that our architectures and training procedures produce fundamental blind spots in our models, and that our models are not learning the same features that humans do.

These adversarial attacks are of interest for more than just the theoretical issues they pose -- concerns have also been raised over the vulnerability of CNNs to these perturbations in the real world, where they are used for mission-critical applications such as online content filtration systems and self-driving cars \cite{Eykholt2017RobustPA, Kurakin2016AdversarialEI}. As a result, a great deal of effort has been dedicated to studying adversarial perturbations. Much of the literature has been dedicated to the development of new attacks that use different perceptibility metrics \cite{Brown2017AdversarialP, Su2017OnePA, Sharif2017AdversarialGN}, security settings (black box/white box) \cite{Papernot2017PracticalBA, Athalye2018ObfuscatedGG}, as well as increasing efficiency \cite{Goodfellow2014ExplainingAH}. Defending against adversarial attacks is also well studied. In particular, adversarial training, where models are trained on adversarial examples, has been shown to be effective under certain assumptions \cite{Madry2017TowardsDL, Sinha2017CertifyingSD}.

%


%

Adversarial attacks can be classified into two categories: white-box attacks and black-box attacks. In white-box attacks, information of the model (i.e., its architecture, gradient information, etc.) is accessible, whereas in black-box attacks, the attackers have access only to the prediction. Black-box attacks are a bigger concern for real-world applications for the obvious reason that such applications typically will not reveal their models publicly, especially when security is a concern (e.g., CNN-based objectionable content filters in social media). Consequently, black-box attacks are mostly focused on the transferability of adversarial examples \cite{Liu2016DelvingIT}.

Moreover, adversarial examples generated using white-box attacks will sometimes successfully attack an unrelated model. This phenomenon is known as ``transferability.'' However, black-box success rates for an attack are nearly always lower than those of white-box attacks, suggesting that the white-box attacks overfit on the source model. Different adversarial attacks transfer at different rates, but most of them are not optimizing specifically for transferability. This paper aims to achieve the goal of increasing the transferability of a given adversarial example. To this end, we propose a novel method that fine-tunes a given adversarial example through examining its representations in intermediate feature maps that we call \textit{Intermediate Level Attack} (ILA).

Our method draws upon two primary intuitions. First, while we do not expect the direction found by the original adversarial attack to be the most optimal for transferability, we do expect it to be a reasonable proxy, as it still transfers far better than random noise would. As such, if we are searching for a more transferable attack, we should be willing to stray from the original attack direction in exchange for increasing the norm\footnote{Perturbations with a higher norm are generally more effective, regardless of layer (holds true for black-box attacks as well).}. However, from the ineffectiveness of random noise on neural networks, we see that straying too far from the original direction will cause a decrease in effectiveness -- even if we are able to increase the norm by a modest amount. Thus, we must balance staying close to the original direction and increasing norm. A natural way to do so is to maximize the projection onto the original adversarial perturbation.

Second, we note that although for transferability we would like to sacrifice some direction in exchange for increasing the norm, we are unable to do so in the image space without changing perceptibility, as norm and perceptibility are intrinsically tied\footnote{Under the standard $\epsilon$-ball constraints.}. However, if we examine the intermediate feature maps, perceptibility (in image space) is no longer intrinsically tied to the norm in an intermediate feature map, and we may be able to increase the norm of the perturbation in that feature space significantly with no change in perceptibility in the image space. We will investigate the effects of perturbing different intermediate feature maps on transferability and provide insights drawn from empirical observations.

Our contributions are as follows:
\begin{itemize}
  \setlength\itemsep{1em}
  \item We propose a novel method, ILA, that enhances black-box adversarial transferability by increasing the perturbation on a pre-specified layer of a model. We conduct a thorough evaluation that shows our method improves upon state-of-the-art methods on multiple models across multiple datasets. See Sec.~\ref{results}.
  \item We introduce a procedure, guided by empirical observations, for selecting a layer that maximizes the transferability using the source model alone, thus obviating the need for evaluation on transfer models during hyperparameter optimization.  See Sec.~\ref{subsec:L}.
  \item Additionally, we provide explanatory insights into the effects of optimizing for adversarial examples using intermediate feature maps. See Sec.~\ref{explanation}.
\end{itemize}

\section{Background and Related Work}\label{background}

\subsection{General Adversarial Attacks}
An adversarial example for a given model is generated by augmenting an image so that in the model's decision space its representation moves into the wrong region. Most prior work in generating adversarial examples for attack focuses on disturbing the softmax output space via the input space \cite{Goodfellow2014ExplainingAH, Madry2017TowardsDL, MoosaviDezfooli2016DeepFoolAS, Dong2017BoostingAA}. Some representative white-box attacks are the following:

\textbf{Gradient Based Approaches} The Fast Gradient Sign Method (FGSM) \cite{Goodfellow2014ExplainingAH} generates an adversarial example with the update rule:
$$ x' = x + \epsilon \cdot sign(\nabla_x J(x,y))$$
It is the linearization of the maximization problem
$$\max_{|x' - x| < \epsilon }J( M(x'), y)$$
where $x$ represents the original image; $x'$ is the adversarial example; $y$ is the ground-truth label; $J$ is the loss function; and $M$ is the model until the final softmax layer. Its iterative version (I-FGSM) applies FGSM iteratively \cite{Kurakin2016AdversarialEI}. Intuitively, this fools the model by increasing its loss, which eventually causes misclassification. In other words, it finds perturbations in the direction of the loss gradient of the last layer (i.e., the softmax layer).

\textbf{Decision Boundary Based Approaches} Deepfool \cite{MoosaviDezfooli2016DeepFoolAS} produces approximately the closest adversarial example iteratively by stepping towards the nearest decision boundary. Universal Adversarial Perturbation \cite{MoosaviDezfooli2017UniversalAP} uses this idea to craft a single image-agnostic perturbation that pushes most of a dataset's images across a model's classification boundary.

\textbf{Model Ensemble Attack} The methods mentioned above are designed to yield the best performance only on the model they are tuned to attack; often, the generated adversarial examples do not transfer to other models.  In contrast, \cite{Liu2016DelvingIT} proposed the Model-based Ensembling Attack that transfers better by avoiding dependence on any specific model. It uses $k$ models with softmax outputs, notated as $J_1$, \ldots, $J_k$, and solves
$$\min_{|x' - x| < \epsilon} - \log \left(\sum_{i = 1}^k \alpha_i J_i(x') 1_y \right) + \lambda d(x, x')$$ 
Using such an approach, the authors showed that the decision boundaries of different CNNs align with each other. Consequently, an adversarial example that fools multiple models is likely to fool other models as well.

\subsection{Intermediate-layer Adversarial Attacks}
A small number of studies have focused on perturbing mid-layer outputs. These include \cite{Mopuri2018GeneralizableDO}, which perturbs mid-layer activations by crafting a single universal perturbation that produces as many spurious mid-layer activations as possible. Another is Feature Adversary Attack \cite{Yuan2017AdversarialEA, Sabour2015AdversarialMO}, which performs a targeted attack by minimizing the distance of the representations of two images in internal neural network layers (instead of in the output layer). However, instead of emphasizing adversarial transferability, it focuses more on internal representations. Results in the paper show that even when given a guide image and a dissimilar target image, it is possible to perturb the target image to produce an embedding similar to that of the guide image.

Two other related works \cite{inkawhich2019feature, rozsa2017lots} focus on perturbing intermediate activation maps for the purpose of increasing adversarial transferability in a method similar to that of \cite{Yuan2017AdversarialEA, Sabour2015AdversarialMO} except they focus on black-box transferability. Their method does not focus on fine-tuning existing adversarial examples and differs significantly in attack methodology from ours.

Another recent work that examines intermediate layers for the purposes of increasing transferability is TAP \cite{Zhou2018TransferableAP}. The TAP attack attempts to maximize the norm between the original image $x$ and the adversarial example $x'$ at all layers. In contrast to our approach, they do not attempt to take advantage of a specific layer's feature representations, instead choosing to maximize the norm of the difference across all layers. In addition, unlike their method which generates an entirely new adversarial example, our method fine-tunes existing adversarial examples, allowing us to leverage existing adversarial attacks.

\section{Approach}
Based on the motivation presented in the introduction, we propose the Intermediate Level Attack (ILA) framework, shown in Algorithm \ref{alg:ILA}. We propose the following two variants, differing in their definition of the loss function $L$. Note that we define $F_l(x)$ as the output at layer $l$ of a network $F$ given an input $x$.

\begin{figure}[htb!]
\begin{algorithmic}[1]
\Require Original image in dataset $x$; Adversarial example $x'$ generated for $x$ by baseline attack; Function $F_l$ that calculates intermediate layer output; $L_\infty$ bound $\epsilon$; Learning rate $lr$; Iterations $n$; Loss function $L$.
\Procedure{ILA}{$x',F_l,\epsilon,lr,L$}
\State $x'' = x$
\State $i = 0$
\While{$i < n$}
\State $\Delta y'_l = F_l(x') - F_l(x)$                             
\State $\Delta y''_l = F_l(x'') - F_l(x)$
\State $x'' = x'' - lr \cdot sign(\nabla_{x''} L(y'_l, y''_l)) $
\State $x'' = clip_{\epsilon}(x'' - x) + x$ 
\State $x'' = clip_{\text{image range}}(x'')$ 
\State $i = i + 1$
\EndWhile
\State \textbf{return} $x''$
\EndProcedure
\end{algorithmic}
\caption{Intermediate Level Attack algorithm}\label{alg:ILA}
\end{figure}

\subsection{Intermediate Level Attack Projection (ILAP) Loss}
Given an adversarial example $x'$ generated by attack method $A$ for natural image $x$, we wish to enhance its transferability by focusing on a layer $l$ of a given network $F$. Although $x'$ is not the optimal direction for transferability, we view $x'$ as a hint for this direction.
We treat $\Delta y'_l = F_l(x') - F_l(x)$ as a directional guide towards becoming more adversarial, with emphasis on the disturbance at layer $l$. Our attack will attempt to find an $x''$ such that $\Delta y''_l = F_l(x'') - F_l(x)$ matches the direction of $\Delta y'_l$ while maximizing the norm of the disturbance in that direction. The high-level idea is that we want to maximize $\text{proj}_{\Delta y'_l} \Delta y''_l$ for the reasons expressed in Section \ref{intro}. Since this is a maximization, we can disregard constants, and this simply becomes the dot product. The objective we solve is given below, and we term it the \textit{ILA projection loss}:
\begin{equation} \label{eq:main_proj}
L(y'_l, y''_l) = - \Delta y''_l \cdot \Delta y'_l
\end{equation}

\subsection{Intermediate Level Attack Flexible (ILAF) Loss}

Since the image $x'$ may not be the optimal direction for us to optimize towards, we may want to give the above loss greater flexibility. We do this by explicitly balancing both norm maximization and also fidelity to the adversarial direction at layer $l$. We note that in a rough sense, ILAF is optimizing for the same thing as ILAP. We augment the above loss by separating out the maintenance of the adversarial direction from the magnitude, and control the trade-off with the additional parameter $\alpha$ to obtain the following loss, termed the \textit{ILA flexible loss}:

\begin{equation} \label{eq:main_flexible}
\begin{gathered}
L(y'_l, y''_l) = \\
- \underbrace{\alpha \cdot \frac{\|\Delta y''_l\|_2}{\|\Delta y'_l\|_2}}_{\text{maximize disturbance}} - \underbrace{\frac{\Delta y''_l}{\|\Delta y''_l\|_2} \cdot \frac{\Delta y'_l}{\|\Delta y'_l\|_2}}_{\text{maintain original direction}}
\end{gathered}
\end{equation}

\subsection{Attack}
In practice, we choose either the ILAP or ILAF loss and iterate $n$ times to attain an approximate solution to the respective maximization objective. Note that the projection loss only has the layer $l$ as a hyperparameter, whereas the flexible loss also has the additional loss weight $\alpha$ as a hyperparameter. The above attack assumes that $x'$ is a pre-generated adversarial example. As such, the attack can be viewed as a fine-tuning of the adversarial example $x'$. We fine-tune for greater norm of the output difference at layer $l$ (which we hope will be conducive to greater transferability) while attempting to preserve the output difference's direction to avoid destroying the original adversarial structure.

\section{Results}
\label{results}
We start by showing that ILAP increases transferability against I-FGSM, MI-FGSM \cite{Dong2017BoostingAA} and Carlini-Wagner \cite{Carlini2017TowardsET} in the context of CIFAR-10 (Sections 4.1 and 4.2). Results for FGSM and Deepfool are shown in Appendix A\footnote{We re-implemented all attacks except Deepfool, for which we used the original publicly provided implementation. For C\&W, we used a randomized targeted version, since it has better performance.}. We test on a variety of models, namely: ResNet18 \cite{He2016DeepRL}, SENet18 \cite{Hu2017SqueezeandExcitationN}, DenseNet121 \cite{Huang2017DenselyCC} and GoogLeNet \cite{Szegedy2015GoingDW}. Architecture details are specified in Appendix A; note that in the below results sections, instead of referring to the architecture specific layer names, we refer to layer indices (e.g. $l=0$ is the last layer of the first block). Our models are trained on CIFAR-10 \cite{Krizhevsky2009LearningML} with the code and hyperparameters in \cite{Liu2018} to final test accuracies of $94.8\%$ for ResNet18, $94.6\%$ for SENet18, $95.6\%$ for DenseNet121, and $94.9\%$ for GoogLeNet.

For a fair comparison, we use the output of an attack $A$ that was run for $20$ iterations as a baseline. ILAP runs for $10$ iterations starting from scratch using the output of attack $A$ after $10$ iterations as the reference adversarial example. The learning rate is set to $0.002$ for both I-FGSM and MI-FGSM\footnote{Tuning the learning rate does not substantially affect transferability, as shown in Appendix G.}.

In Section 4.2 we also show that we can select a nearly-optimal layer for transferability using only the source model. Moreover, ILAF allows further tuning to improve the performance across layers (Section 4.3).

Finally, we demonstrate that ILAP also improves transferability under the more complex setting of ImageNet \cite{Deng2009ImageNetAL} and that it supercedes state-of-the-art attacks focused on increasing transferability, namely the Zhou et al. attack (TAP) \cite{Zhou2018TransferableAP} and the Xie et al. attack \cite{Xie_2019_CVPR} (Section 4.4).

\subsection{ILAP Targeted at Different $L$ Values}

\begin{figure}[htb]
    \centering
    \includegraphics[width=0.45\textwidth]{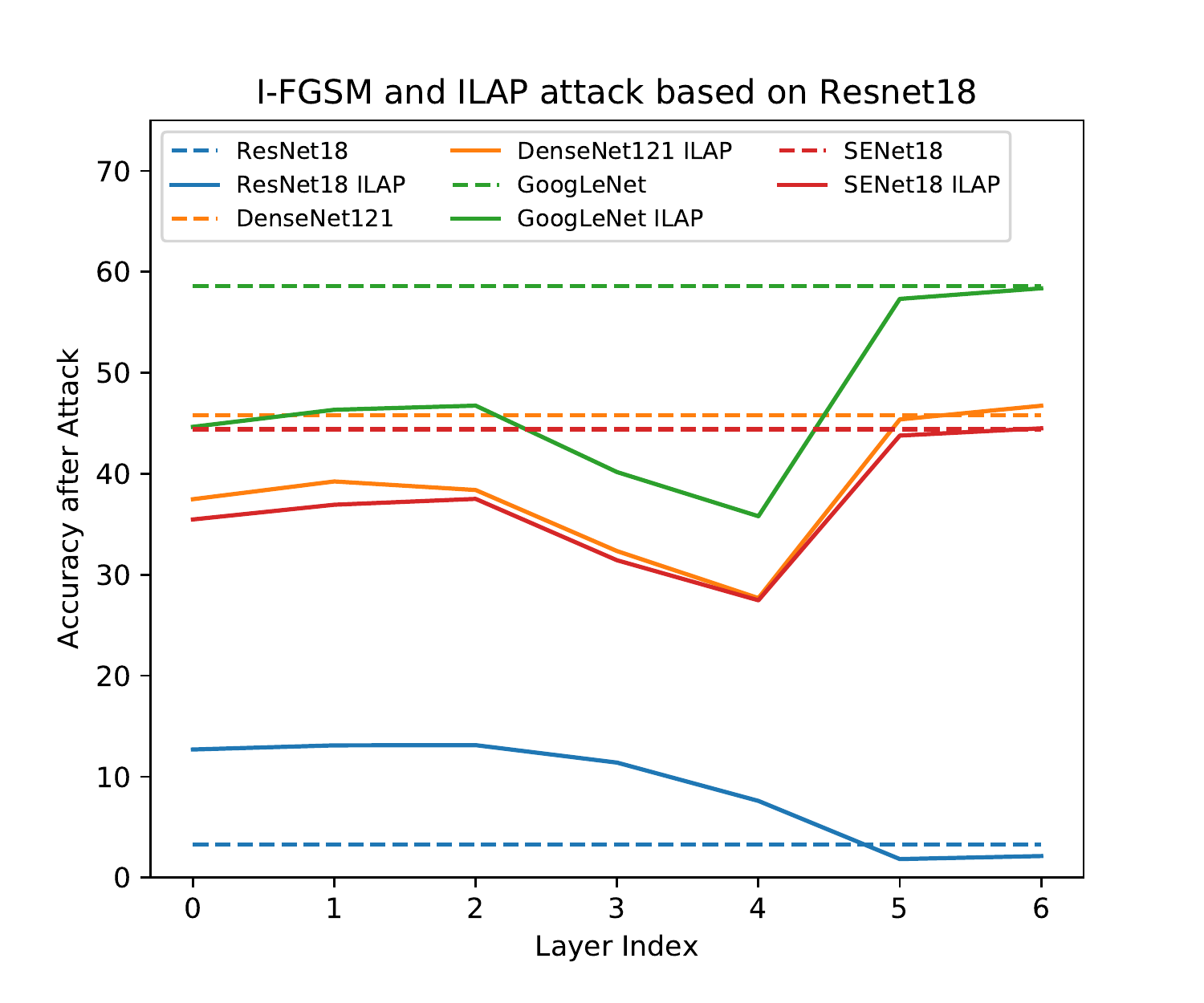}
    \caption{\label{fig:manymodeltransfer} Transfer results of ILAP against I-FGSM on ResNet18 as measured by DenseNet121, SENet18, and GoogLeNet on CIFAR-10 (lower accuracies indicate better attack). }
\end{figure}

To confirm the effectiveness of our attack, we fix a single source model and baseline attack method, and then check how ILAP transfers to the other models compared to the baseline attack. Results for ResNet18 as the source model and I-FGSM as the baseline method are shown in Figure \ref{fig:manymodeltransfer}. Comparing the results of both methods on the other models, we see that ILAP outperforms I-FGSM when targeting any given intermediate layer, and does especially well for the optimal hyperparameter value of $l = 4$. Note that the choice of layer is important for both performance on the source model and target models.
Full results are shown in Appendix A.

\begin{figure}[htb]
    \centering
    \includegraphics[width=0.4\textwidth]{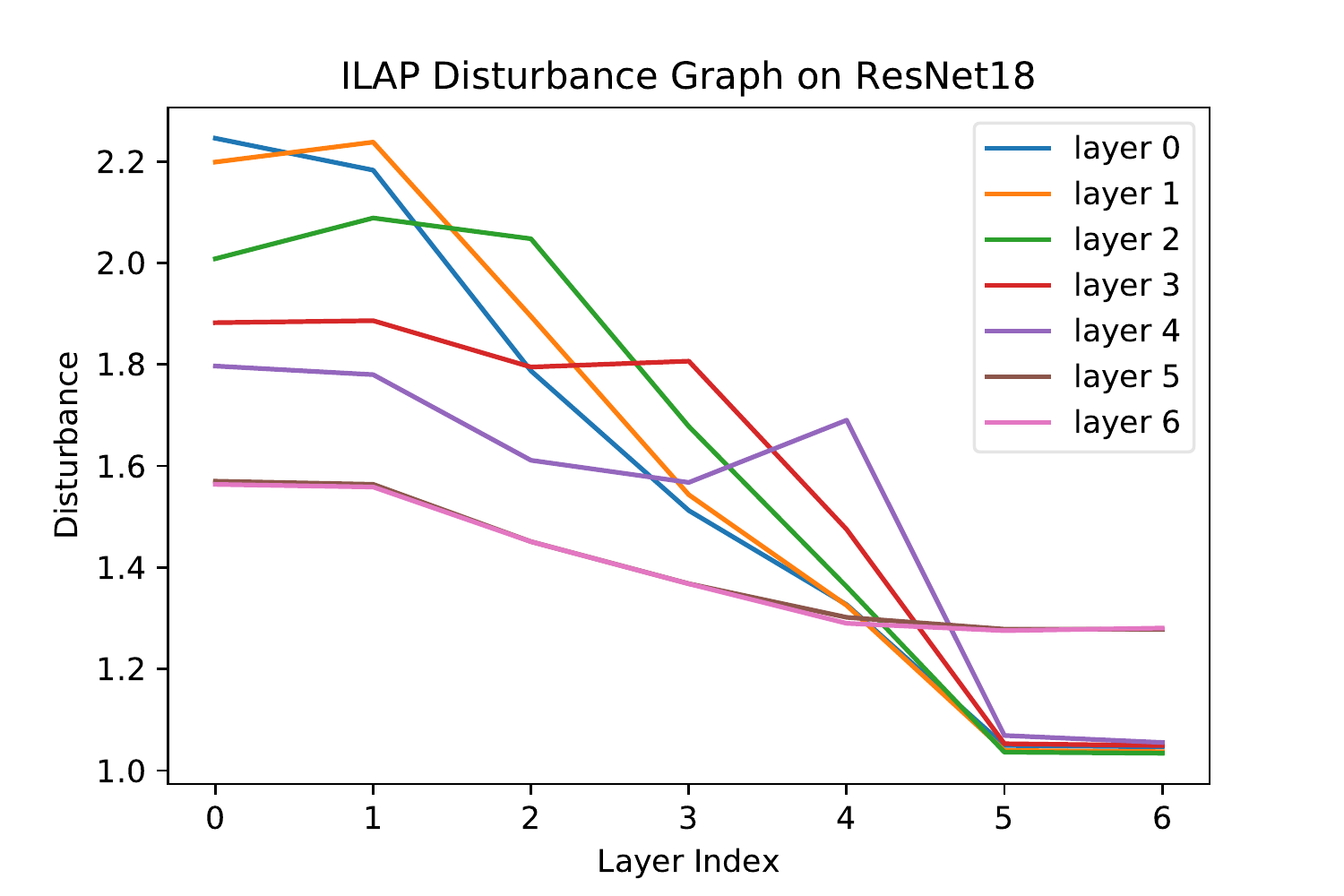}
    \caption{\label{fig:peaks} Disturbance values $\left(\frac{\|\Delta y''_l\|_2}{\|\Delta y'_l\|_2}\right)$ at each layer for ILAP targeted at layer $l = 0,1,...,6$ for ResNet18. Observe that the $l$ in the legend refers to the hyperparameter set in the ILAP attack, and afterwards the disturbance values were computed on layers indicated by the $l$ in the x-axis. Note that the last peak is produced by the $l = 4$ ILAP attack.}
\end{figure}


\begin{table}[!htb]
\caption{\label{tab:ilap-vs-sota} ILAP vs. State-of-the-art Transfer Attacks}
\centering
\begin{threeparttable}
\begin{tabular}{ccccc}
\toprule
    & \multicolumn{2}{c}{TAP \cite{Zhou2018TransferableAP}} & \multicolumn{2}{c}{DI$^2$-FGSM \cite{Xie_2019_CVPR}} \\
    \cmidrule(r){2-3}
    \cmidrule(r){4-5}
    Transfer & 20 Itr & Opt ILAP & 20 Itr & Opt ILAP \\
\midrule
  Inc-v4 & 21.5\% & \textbf{16.3\%} & 50.2\% & \textbf{26.7\%} \\
  IncRes-v2 & 26.0\% & \textbf{21.7\%} & 54.6\% & \textbf{29.3\%} \\
\bottomrule
\end{tabular}
\begin{tablenotes}
  \item[$\dagger$] Same model as source model.
  \end{tablenotes}
\end{threeparttable}
\caption*{Table \ref{tab:ilap-vs-sota}. Same as experiment in Table \ref{tab:main-result} but with TAP and DI$^2$-FGSM from Xie et al. \cite{Xie_2019_CVPR}. Evaluation is performed with 5000 randomly selected ImageNet validation set images, and $\epsilon = 0.06$. The source model used is Inc-v3 and the target layer specified for ILAP is Conv2d\_4a\_3x3. }
\end{table}

\subsection{ILAP with Pre-Determined $L$ Value}
\label{subsec:L}

Above we demonstrated that adversarial examples produced by ILAP exhibit the strongest transferability when targeting a specific layer (i.e. choosing a layer as the $l$ hyperparameter). We wish to pre-determine this optimal value based on the source model alone, so as to avoid tuning the hyperparameter $l$ by evaluating on other models. To do this, we examine the relationship between transferability and the ILAP layer disturbance values for a given ILAP attack. We define the disturbance values of an ILAP attack perturbation $x''$ as values of the function $f(l) = \frac{\|\Delta y''_l\|_2}{\|\Delta y'_l\|_2}$ for all values of $l$ in the source model. For each value of $l$ in ResNet18 (the set of $l$ is defined for each architecture in Appendix A) we plot the disturbance values of the corresponding ILAP attack in Figure \ref{fig:peaks}. The same figure is given for other models in Appendix B. 

We notice that the adversarial examples that produce the latest peak in the graph are typically the ones that have highest transferability for all transferred models (Table \ref{tab:main-result}). Given this observation, we propose that the latest $l$ that still exhibits a peak is a nearly optimal value of $l$ (in terms of maximizing transferability). For example, according to Figure \ref{fig:peaks}, we would choose $l = 4$ as the expected optimal hyperparameter for ILAP with ResNet18 as the source model. Table \ref{tab:main-result} supports our claim and shows that selecting this layer gives an optimal or near-optimal attack. We discuss our discovered explanatory insights for this method in Section \ref{subsec:expl-final}.

\setlength{\tabcolsep}{4pt}
\begin{table*}[!htb]
  \caption{\label{tab:main-result} ILAP Results}
  \centering
  \begin{threeparttable}
   \begin{tabular}{cccccccc}
    \toprule
    & & \multicolumn{3}{c}{MI-FGSM} & \multicolumn{3}{c}{C \& W} \\
    \cmidrule(r){3-5} \cmidrule(r){6-8}
    Source & Transfer & 20 Itr & 10 Itr ILAP & Opt ILAP  & 1000 Itr & 500 Itr ILAP &  Opt ILAP \\
    \midrule
    & $\text{ResNet18}^\dagger$ & 5.7\% & 11.3\% & \textbf{2.3\%} (6)& 7.3\% &             5.2\% &             \textbf{2.1\%} (5) \\
    ResNet18 & SENet18 & 33.8\% & \textbf{30.6\%} & \textbf{30.6\%} (4)& 85.4\% &            \textbf{41.7\%} &            \textbf{41.7\%} (4) \\
    ($l=4$) & DenseNet121 & 35.1\% & \textbf{30.4\%} & \textbf{30.4\%} (4)&84.4\% &            \textbf{41.7\%} &            \textbf{41.7\%} (4) \\
    & GoogLeNet & 45.1\% & \textbf{37.7\%} & \textbf{37.7\%} (4)& 90.6\% &            \textbf{57.3\%} &            \textbf{57.3\%} (4) \\
    \midrule
    
    & ResNet18 & 31.0\% & \textbf{27.5\%} & \textbf{27.5\%} (4)& 87.5\% &            \textbf{42.7\%} &            \textbf{42.7\%} (4) \\
    SENet18 & $\text{SENet18}^\dagger$ & 3.3\% & 10.0\% & \textbf{2.6\%} (6)& 6.2\% &             7.3\% &             \textbf{3.1\%} (5) \\
    ($l=4$) & DenseNet121 & 31.6\% & \textbf{27.3\%}& \textbf{27.3\%} (4)& 88.5\% &            \textbf{38.5\%} &            \textbf{38.5\%} (4) \\
    & GoogLeNet & 41.1\% & \textbf{34.8\%} &\textbf{34.8\%} (4)& 91.7\% &            \textbf{52.1\%} &            \textbf{52.1\%} (4) \\
    \midrule
    
    & ResNet18 & 34.4\% & \textbf{28.1}\% & \textbf{28.1\%}(6)& 87.5\% &            \textbf{37.5\%} &            \textbf{37.5\%} (6) \\
    DenseNet121 & SENet18 & 33.5\% & \textbf{27.7\%} & \textbf{27.7\%} (6)& 86.5\% &            \textbf{34.4\%} &            \textbf{34.4\%} (6) \\
    ($l=6$) & $\text{DenseNet121}^\dagger$ & 6.4\% & 4.0\% & \textbf{0.8\%}(9)& 2.1\% &             \textbf{0.0\%} &             \textbf{0.0\%} (9) \\
    & GoogLeNet & 36.3\% & \textbf{30.3\%} & \textbf{30.3\%} (6)& 90.6\% &            \textbf{45.8\%} &            \textbf{45.8\%} (6) \\
    \midrule
    
    & ResNet18 & 44.6\% & 34.5\% &\textbf{33.2\%}(3)& 89.6\% &            63.5\% &            \textbf{60.4\%} (7) \\
    GoogLeNet & SENet18 & 43.0\% & 33.5\% & \textbf{32.6\%}(3)&  90.6\% &            \textbf{53.1\%} &            \textbf{53.1\%} (9) \\
    ($l=9$) & DenseNet121 & 38.9\% & 29.2\% & \textbf{28.8\%}(3)& 89.6\% &            58.3\% &            \textbf{51.0\%} (8) \\
    & $\text{GoogLeNet}^\dagger$ & 1.5\% & 1.4\% & \textbf{0.5\%} (11)& 4.2\% &             \textbf{0.0\%} &            \textbf{0.0\%} (12) \\
    \bottomrule
  \end{tabular}
  \begin{tablenotes}
  \item[$\dagger$] Same model as the source model.
  \end{tablenotes}
  \end{threeparttable}
\caption*{Table \ref{tab:main-result}. Accuracies after attack are shown for the models (lower accuracies indicate better attack). The hyperparameter $l$ in the ILAP attack is being fixed for each source model as decided by the layer disturbance graphs (e.g. setting $l=4$ for ResNet18 since it was the last peak in Figure \ref{fig:peaks}).  ``Opt ILAP'' refers to a 10 iteration ILAP that chooses the optimal layer (determined by evaluating on transfer models). Perhaps surprisingly, ILAP beats out the baseline attack on the original model as well.}
\end{table*}

\subsection{ILAF vs. ILAP}
We show that ILAF can further improve transferability with the additional tunable hyperparameter $\alpha$. The best ILAF result for each model improves over ILAP as shown in Table \ref{tab:ilap-vs-ilaf}. However, note that the optimal $\alpha$ differs for each model and requires substantial hyperparameter tuning to outperform ILAP. Thus, ILAF can be seen as a more model-specific version that requires more tuning, whereas ILAP works well out of the box. Full results are in Appendix C.
\begin{table}[!tbh]
  \caption{\label{tab:ilap-vs-ilaf} ILAP vs. ILAF}
  \centering
  \begin{tabular}{ccc}
    \toprule
    Model &  ILAP (best) & ILAF (best) \\
    \midrule
    DenseNet121 & 27.7\% & 26.6\% \\
    GoogLeNet  & 35.8\% & 34.7\% \\
    SENet18  & 27.5\% & 26.3\% \\
    \bottomrule
  \end{tabular}
\caption*{Table \ref{tab:ilap-vs-ilaf}. Here we show the difference in transfer performance between ILAP vs. ILAF generated using ResNet18 (with optimal hyperparameters for both attacks). }
\end{table}



\subsection{ILAP on ImageNet}

We also tested ILAP on ImageNet, with ResNet18, DenseNet121, SqueezeNet, and AlexNet pretrained on ImageNet (as provided in \cite{Marcel2010TorchvisionTM}). The learning rates for all attacks are tuned for best performance. For I-FGSM the learning rate is set to $0.008$, for ILAP with I-FGSM to $0.01$, for MI-FGSM to $0.018$, and for ILAP with MI-FGSM to $0.018$. To evaluate transferability, we tested the accuracies of different models over adversarial examples generated from all $50000$ ImageNet test images. We observe that ILAP improves over I-FGSM and MI-FGSM on ImageNet. Results for ResNet18 as the source model and I-FGSM as the baseline attack are shown in Figure \ref{fig:imagenet}. Full results are in Appendix D.

In order to show our approach outperforms pre-existing methods, we tested ILAP against both TAP \cite{Zhou2018TransferableAP}\footnote{Code was not made available for this paper, hence we reproduced their method to the best of our ability.} and Xie et al. \cite{Xie_2019_CVPR}\footnote{Pretrained ImageNet models for Inc-v3, Inc-v4, and IncRes-v2 were obtained from Cadene's Github repo \cite{Cadene2019}.} in an ImageNet setting. The results are shown in Table \ref{tab:ilap-vs-sota}\footnote{Previous versions of this paper included results with incorrect normalization. Results indicating that ILAP is competitive with TAP on CIFAR-10 are in Appendix H.}.
\begin{figure}[htb]
    \centering
    \includegraphics[width=0.45\textwidth]{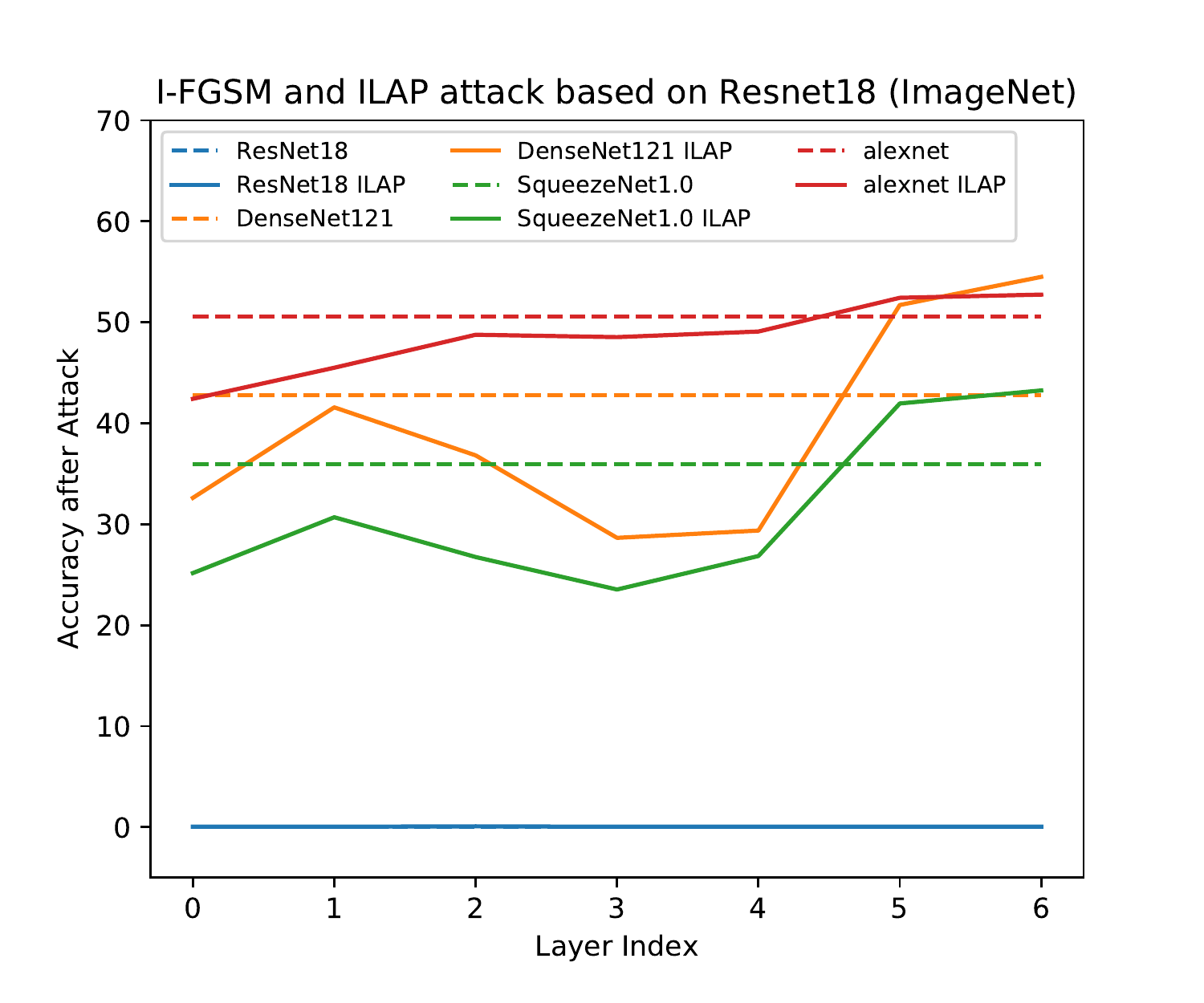}
    \caption{\label{fig:imagenet} Transfer results of ILAP against I-FGSM on ResNet18 as measured by DenseNet121, SqueezeNet, and AlexNet on ImageNet (lower accuracies indicate better attack).}
\end{figure}

\section{Explaining the Effectiveness of Intermediate Layer Emphasis}
\label{explanation}


At a high level, we motivated projection in an intermediate feature map as a way to increase transferability. We saw empirically that it was desireable to target the layer corresponding to the latest peak (see Figure \ref{fig:peaks}) on the source model in order to maximize transferability. In this section, we attempt to explain the factors causing ILAP performance to vary across layers as well as what they suggest about the optimal layer for ILAP. As we iterate through layer indices, there are two factors affecting our performance: the angle between the original perturbation direction and best transfer direction (defined below in Section \ref{subsec:expl-angle}) as well as the linearity of the model decision boundary.

Below, we discuss how the factors change across layers and affect transferability of our attack.

\subsection{Angle between the Best Transfer Direction and the Original Perturbation}
\label{subsec:expl-angle}

Motivated by \cite{Liu2016DelvingIT} (where it is shown that the decision boundaries of models with different architectures often align) we define the \textit{Best Transfer Direction} (BTD):

\textbf{Best Transfer Direction}: Let $x$ be an image and $M$ be a large (but finite) set of distinct CNNs. Find $x'$ such that 
\[
x' = \argmax_{x'\:s.t.\:|x'-x| < \epsilon} \sum_{m\in M} \mathbbold{1}[m(x') \neq m(x)]
\]
Then the \textit{Best Transfer Direction} of x is $BTD_x = \frac{x' - x}{\|x'- x\|}$.


Since our method uses the original perturbation as an approximation for the BTD, it is intuitive that the better this approximation is in the current feature representation, the better our attack will perform. 

\begin{figure}[htb]
    \centering
    \includegraphics[width=0.45\textwidth]{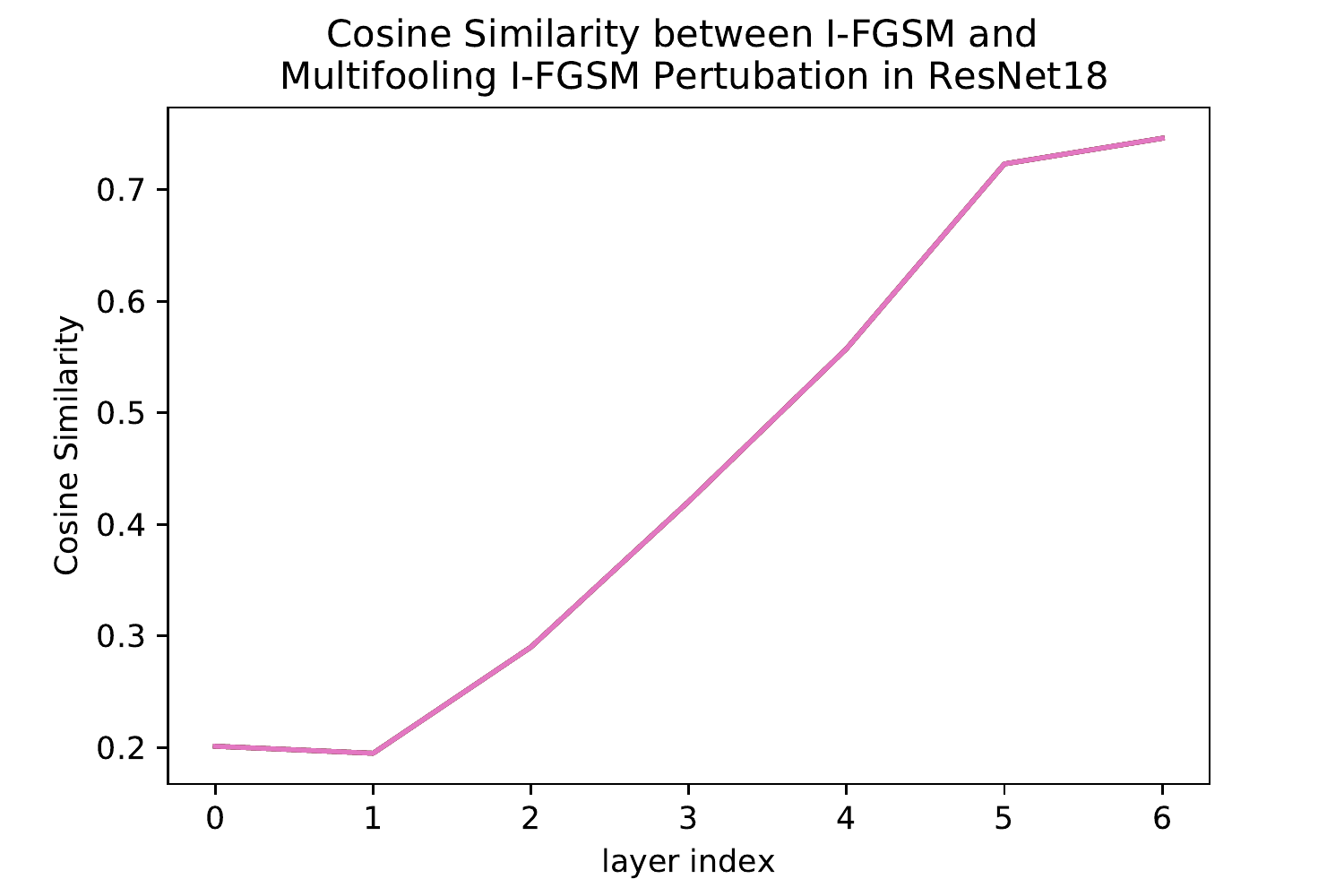}
    \caption{\label{fig:angledifference} As shown in the above figure, in terms of angle, I-FGSM produces a better approximation for the estimated best transfer direction as we increase the layer index. }
\end{figure}

We want to investigate the nature of how well a chosen source model attack, like I-FGSM, aligns with the BTD throughout layers. Here we measure alignment between an I-FGSM perturbation and an empirical estimate of the BTD (a multi-fool perturbation of the four models we evaluate on in the CIFAR-10 setting) using the angle between them. We investigate the alignment between the feature map outputs of the I-FGSM perturbation and the BTD at each layer. As shown in Figure \ref{fig:angledifference}, the angle between the perturbation of I-FGSM and that of the BTD decreases as we iterate the layer indices. Therefore, the later the target layer is in the source model, the better it is to use I-FGSM's attack direction as a guide. This is a factor \textit{increasing} transfer attack success rate as layer indices increase.

\begin{figure}[htb]
    \centering
    \includegraphics[width=0.45\textwidth]{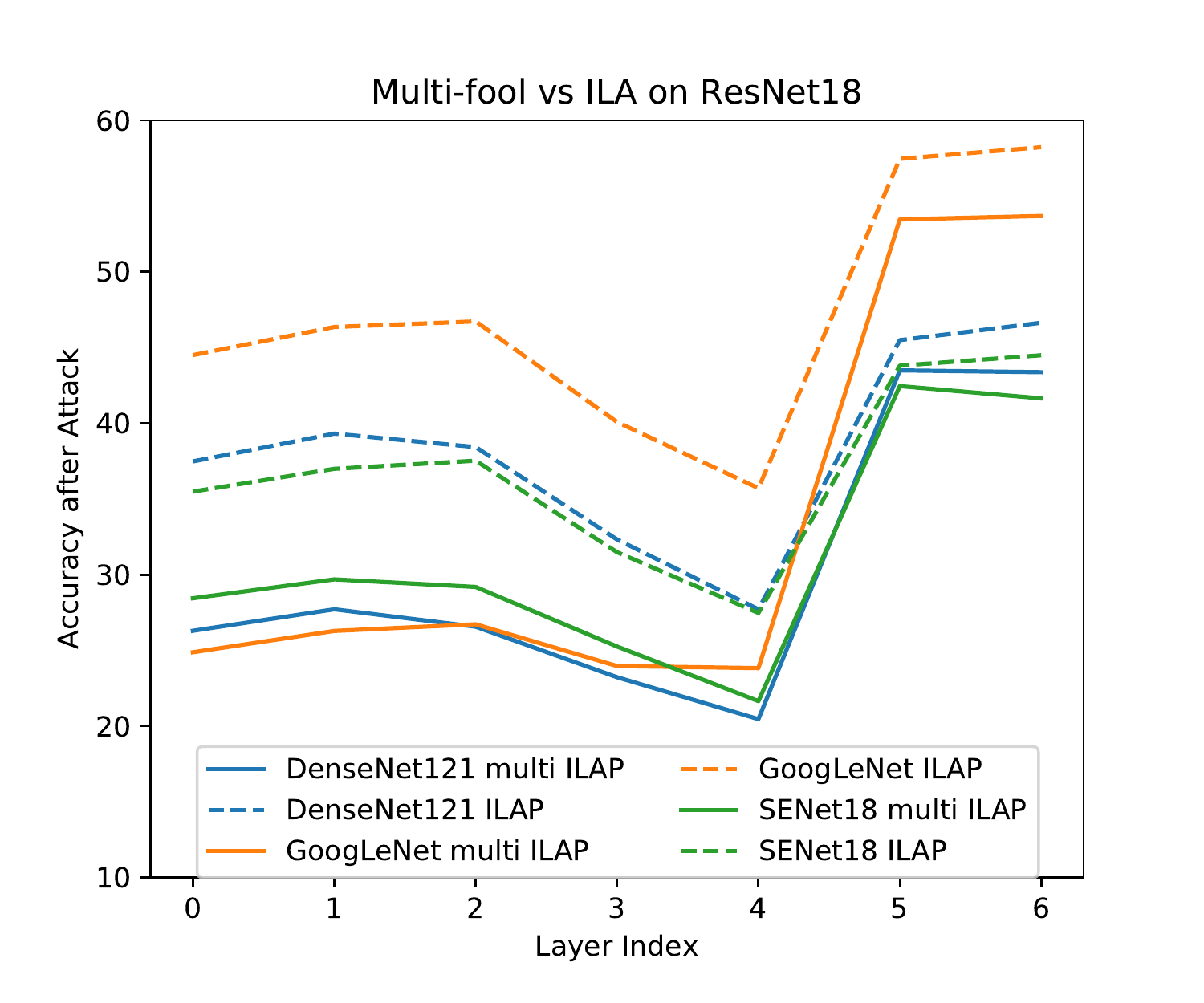}
    \caption{\label{fig:resnetmultifool} Here we show that ILAP with a better approximation for BTD (multi-fool) performs better. In addition, using a better approximation for BTD disproportionately improves the earlier layers' performance. }
\end{figure}

To test our hypothesis, we propose to eliminate this source of variation in performance by using a multi-fool perturbation as the starting perturbation for ILAP, which is a better approximation for the BTD. As shown in Figure \ref{fig:resnetmultifool}, ILAP performs substantially better when using a multi-fool perturbation as a guide rather than an I-FGSM perturbation, thus confirming that using a better approximation of the BTD gives better performance for ILAP. In addition, we see that these results correspond with what we would expect from Figure \ref{fig:angledifference}. In the earlier layers, I-FGSM is a worse approximation of the BTD, so passing in a multi-fool perturbation improves performance significantly. In the later layers, I-FGSM is a much better approximation of the BTD, and we see that passing in a multi-fool perturbation does not increase performance much.

\subsection{Linearity of Decision Boundary}
\label{subsec:expl-linearity}
If we view I-FGSM as optimizing to cross the decision boundary, we can interpret ILAP as optimizing to cross the decision boundary approximated with a hyper-plane perpendicular to the I-FGSM perturbation. As the layer indices increase, the function from the feature space to the final output of the source model tends to becomes increasingly linear (there are more nonlinearities between earlier layers and the final layer than there are between a later layer and the final layer). In fact, we note that at the final layer, the decision boundary is completely linear. 
Thus, our linear approximation of the decision boundary becoming more accurate is one factor in improving ILAP performance as we select the later layers.

We define the ``true decision boundary" as a majority-vote ensemble of a large number of CNNs. Note that for transfer, we care less about how well we are approximating the source model decision boundary than we do about how well we are approximating the true decision boundary. In most feature representations we expect that the true decision boundary is more linear, as ensembling reduces variance. However, note that at least in the final layer, by virtue of the source model decision boundary being exactly linear, the true decision boundary cannot be more linear, and is likely to be less linear.

\begin{figure}[htb]
    \centering
    \includegraphics[width=0.45\textwidth]{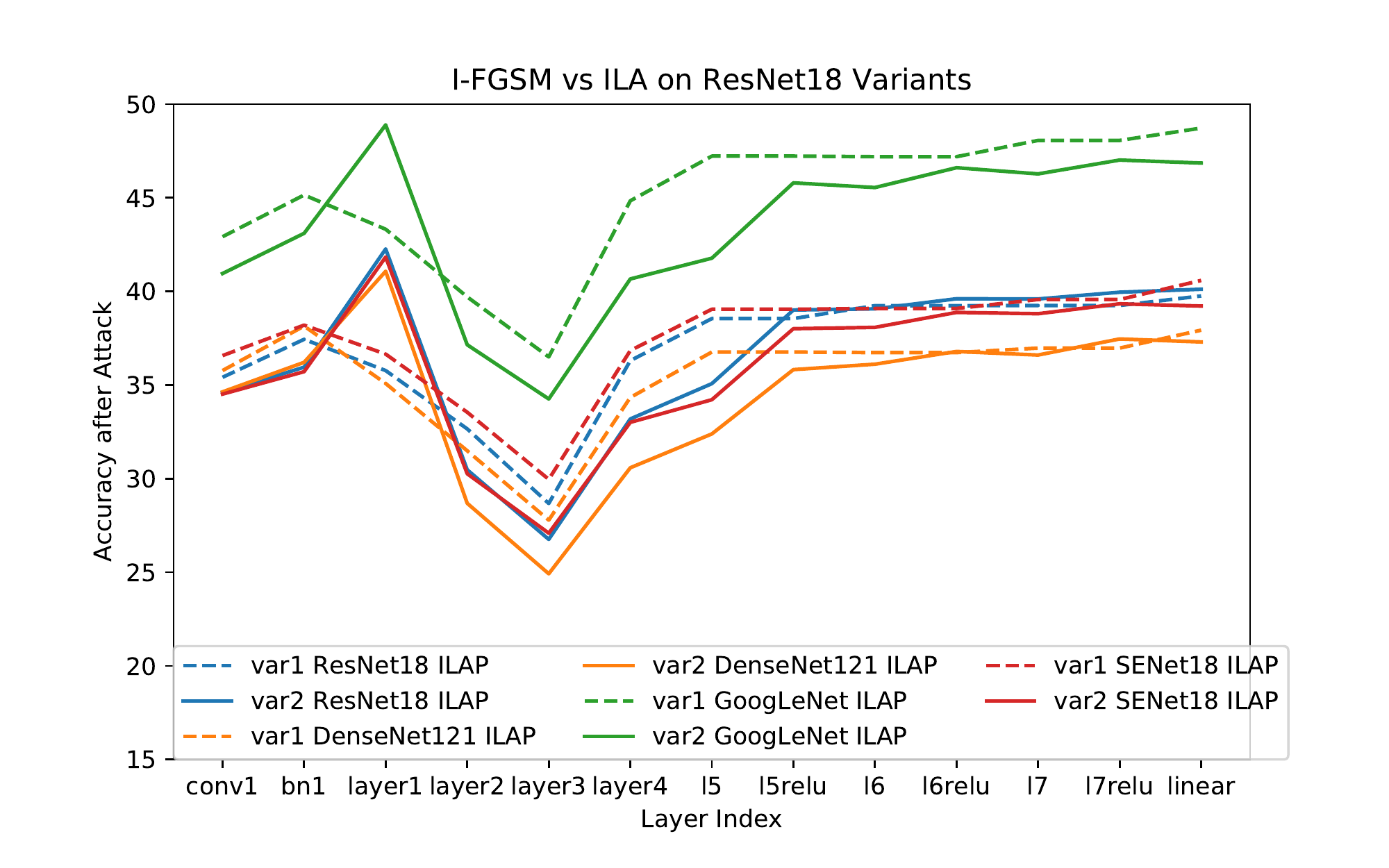}
    \caption{\label{fig:resnetlinear} When there is more nonlinearity present in the later portion of the network, the performance of ILAP does not deteriorate as rapidly. Variant 1 (var1) is the version of ResNet18 with additional linear layers not followed by activations, while Variant 2 (var2) does have activations.}
\end{figure}

We hypothesize that this flip is what causes us to perform worse in the final layers. In these layers, the source model decision boundary is more linear than the true decision boundary, so our approximation performs poorly. We test this hypothesis by attacking two variants of ResNet18 augmented with 3 linear layers before the last layer: one variant without activations following the added layers (var1) and one with (var2). As shown in Figure \ref{fig:resnetlinear}, ILAP performance decreases less in the second variant. Also note that these nonlinearities also cause worse ILAP performance earlier in the network.

Thus, we conclude that the extreme linearity of the last several layers is associated with ILAP performing poorly.

\subsection{Explanation of the main result}
\label{subsec:expl-final}
In this section, we tie together all of the above factors to explain the optimal intermediate layer for transferability. Denote:

\begin{itemize}
\item the decreasing angle difference between I-FGSM's and BTD's perturbation direction as Factor 1
\item the increasing linearity with respect to the decision boundary as we increase layer index as Factor 2, and
\item the excessive linearity of the source model decision boundary as Factor 3
\end{itemize}

On the transfer models, as the index of the attacked source model layer increases, Factors 1 and 2 increase attack rate, while Factor 3 decreases the attack rate. Thus, before some layer, Factors 1 and 2 cause transferability to increase as layer index increases; however, afterward, Factor 3 wins out and causes transferability to decrease as the layer index increases. Thus the layer right before the point where this switch happens is the layer that is optimal for transferability.

We note that this explanation would also justify the method presented in Section \ref{subsec:L}. Intuitively, having a peak corresponds with having the linearized decision boundary (from using projection as the objective) be very different from the source model's decision boundary. If this were not the case, then I-FGSM would presumably have found this improved perturbation already. As such, choosing the last layer that we can get a peak at corresponds with both having as linear of a decision boundary as possible (as late of a layer as possible) while still having enough room to move (the peak).

On the source model, since there is no notion of a ``transfer" attack, Factor 3 and Factor 1 do not have any effect. Therefore, Factor 2 causes the performance of the later layers to improve, so much so that at the final layer ILAP's performance on the source model is actually equal or better on all the attacks we used as baselines (see Figure \ref{fig:manymodeltransfer}). We hypothesize the improved performance on the source model is the result of a simpler loss and thus an easier to optimize loss landscape.

\section{Conclusion}
We introduce a novel attack, coined ILA, that aims to enhance the transferability of any given adversarial example. It is a framework with the goal of enhancing transferability by increasing projection onto the \textit{Best Transfer Direction}. Within this framework, we propose two variants, ILAP and ILAF, and analyze their performance. We demonstrate that there exist specific intermediate layers that we can target with ILA to substantially increase transferability with respect to the attack baselines. In addition, we show that a near-optimal target layer can be selected without any knowledge of transfer performance. Finally, we provide some intuition regarding ILA's performance and why it performs differently in different feature spaces.

Potential future works include making use of the interactions between ILA and existing adversarial attacks to explain differences among existing attacks, as well as extending ILA to perturbations produced for different settings (universal or targeted perturbations). In addition, other methods of attacking intermediate feature spaces could be explored, taking advantage of the properties we explored in this paper.

\subsection*{Acknowledgements}
We want to thank Pian Pawakapan, Prof. Kavita Bala, and Prof. Bharath Hariharan for helpful discussions. This work is supported in part by a Facebook equipment donation.

{\small
\bibliographystyle{ieee_fullname}
\bibliography{ila}

\begin{thebibliography}{10}\itemsep=-1pt

\bibitem{Athalye2018ObfuscatedGG}
Anish Athalye, Nicholas Carlini, and David~A. Wagner.
\newblock Obfuscated gradients give a false sense of security: Circumventing
  defenses to adversarial examples.
\newblock In {\em ICML}, 2018.

\bibitem{Brown2017AdversarialP}
Tom~B. Brown, Dandelion Man{\'e}, Aurko Roy, Mart{\'i}n Abadi, and Justin
  Gilmer.
\newblock Adversarial patch.
\newblock {\em CoRR}, abs/1712.09665, 2017.

\bibitem{Cadene2019}
Remi Cadene.
\newblock pretrained-models.pytorch.
\newblock \url{https://github.com/Cadene/pretrained-models.pytorch}, 2019.

\bibitem{Carlini2017TowardsET}
Nicholas Carlini and David~A. Wagner.
\newblock Towards evaluating the robustness of neural networks.
\newblock {\em 2017 IEEE Symposium on Security and Privacy (SP)}, pages 39--57,
  2017.

\bibitem{Deng2009ImageNetAL}
Jia Deng, Wei Dong, Richard Socher, Li-Jia Li, Kai Li, and Li Fei-Fei.
\newblock Imagenet: A large-scale hierarchical image database.
\newblock {\em 2009 IEEE Conference on Computer Vision and Pattern
  Recognition}, pages 248--255, 2009.

\bibitem{Dong2017BoostingAA}
Yinpeng Dong, Fangzhou Liao, Tianyu Pang, Hang Su, Jun Zhu, Xiaolin Hu, and
  Jianguo Li.
\newblock Boosting adversarial attacks with momentum.
\newblock 2017.

\bibitem{Eykholt2017RobustPA}
Kevin Eykholt, Ivan Evtimov, Earlence Fernandes, Bo Li, Amir Rahmati, Chaowei
  Xiao, Atul Prakash, Tadayoshi Kohno, and D. Song.
\newblock Robust physical-world attacks on deep learning models.
\newblock 2017.

\bibitem{Goodfellow2014ExplainingAH}
Ian~J. Goodfellow, Jonathon Shlens, and Christian Szegedy.
\newblock Explaining and harnessing adversarial examples.
\newblock {\em CoRR}, abs/1412.6572, 2014.

\bibitem{He2016DeepRL}
Kaiming He, Xiangyu Zhang, Shaoqing Ren, and Jian Sun.
\newblock Deep residual learning for image recognition.
\newblock {\em 2016 IEEE Conference on Computer Vision and Pattern Recognition
  (CVPR)}, pages 770--778, 2016.

\bibitem{Hu2017SqueezeandExcitationN}
Jie Hu, Li Shen, and Gang Sun.
\newblock Squeeze-and-excitation networks.
\newblock {\em CoRR}, abs/1709.01507, 2017.

\bibitem{Huang2017DenselyCC}
Gao Huang, Zhuang Liu, Laurens van~der Maaten, and Kilian~Q. Weinberger.
\newblock Densely connected convolutional networks.
\newblock {\em 2017 IEEE Conference on Computer Vision and Pattern Recognition
  (CVPR)}, pages 2261--2269, 2017.

\bibitem{inkawhich2019feature}
Nathan Inkawhich, Wei Wen, Hai~Helen Li, and Yiran Chen.
\newblock Feature space perturbations yield more transferable adversarial
  examples.
\newblock In {\em Proceedings of the IEEE Conference on Computer Vision and
  Pattern Recognition}, pages 7066--7074, 2019.

\bibitem{Krizhevsky2009LearningML}
Alex Krizhevsky.
\newblock Learning multiple layers of features from tiny images.
\newblock 2009.

\bibitem{Krizhevsky2012ImageNetCW}
Alex Krizhevsky, Ilya Sutskever, and Geoffrey~E. Hinton.
\newblock Imagenet classification with deep convolutional neural networks.
\newblock In {\em NIPS}, 2012.

\bibitem{Kurakin2016AdversarialEI}
Alexey Kurakin, Ian~J. Goodfellow, and Samy Bengio.
\newblock Adversarial examples in the physical world.
\newblock {\em CoRR}, abs/1607.02533, 2016.

\bibitem{Liu2018}
Kuang Liu.
\newblock Pytorch cifar10.
\newblock \url{https://github.com/kuangliu/pytorch-cifar}, 2018.

\bibitem{Liu2016DelvingIT}
Yanpei Liu, Xinyun Chen, Chang Liu, and Dawn~Xiaodong Song.
\newblock Delving into transferable adversarial examples and black-box attacks.
\newblock {\em CoRR}, abs/1611.02770, 2016.

\bibitem{Madry2017TowardsDL}
Aleksander Madry, Aleksandar Makelov, Ludwig Schmidt, Dimitris Tsipras, and
  Adrian Vladu.
\newblock Towards deep learning models resistant to adversarial attacks.
\newblock {\em CoRR}, abs/1706.06083, 2017.

\bibitem{Marcel2010TorchvisionTM}
S{\'e}bastien Marcel and Yann Rodriguez.
\newblock Torchvision the machine-vision package of torch.
\newblock In {\em ACM Multimedia}, 2010.

\bibitem{MoosaviDezfooli2017UniversalAP}
Seyed-Mohsen Moosavi-Dezfooli, Alhussein Fawzi, Omar Fawzi, and Pascal
  Frossard.
\newblock Universal adversarial perturbations.
\newblock {\em 2017 IEEE Conference on Computer Vision and Pattern Recognition
  (CVPR)}, pages 86--94, 2017.

\bibitem{MoosaviDezfooli2016DeepFoolAS}
Seyed-Mohsen Moosavi-Dezfooli, Alhussein Fawzi, and Pascal Frossard.
\newblock Deepfool: A simple and accurate method to fool deep neural networks.
\newblock {\em 2016 IEEE Conference on Computer Vision and Pattern Recognition
  (CVPR)}, pages 2574--2582, 2016.

\bibitem{Mopuri2018GeneralizableDO}
Konda~Reddy Mopuri, Aditya Ganeshan, and R.~Venkatesh Babu.
\newblock Generalizable data-free objective for crafting universal adversarial
  perturbations.
\newblock {\em IEEE transactions on pattern analysis and machine intelligence},
  2018.

\bibitem{Papernot2017PracticalBA}
Nicolas Papernot, Patrick~D. McDaniel, Ian~J. Goodfellow, Somesh Jha, Z.~Berkay
  Celik, and Ananthram Swami.
\newblock Practical black-box attacks against machine learning.
\newblock In {\em AsiaCCS}, 2017.

\bibitem{rozsa2017lots}
Andras Rozsa, Manuel G\"unther, and Terrance~E. Boult.
\newblock {LOTS} about attacking deep features.
\newblock In {\em International Joint Conference on Biometrics (IJCB)}, 2017.

\bibitem{Sabour2015AdversarialMO}
Sara Sabour, Yanshuai Cao, Fartash Faghri, and David~J. Fleet.
\newblock Adversarial manipulation of deep representations.
\newblock {\em CoRR}, abs/1511.05122, 2015.

\bibitem{Sharif2017AdversarialGN}
Mahmood Sharif, Sruti Bhagavatula, Lujo Bauer, and Michael~K. Reiter.
\newblock Adversarial generative nets: Neural network attacks on
  state-of-the-art face recognition.
\newblock {\em CoRR}, abs/1801.00349, 2017.

\bibitem{Sinha2017CertifyingSD}
Aman Sinha, Hongseok Namkoong, and John~C. Duchi.
\newblock Certifying some distributional robustness with principled adversarial
  training.
\newblock 2017.

\bibitem{Su2017OnePA}
Jiawei Su, Danilo~Vasconcellos Vargas, and Kouichi Sakurai.
\newblock One pixel attack for fooling deep neural networks.
\newblock {\em CoRR}, abs/1710.08864, 2017.

\bibitem{Szegedy2015GoingDW}
Christian Szegedy, Wei Liu, Yangqing Jia, Pierre Sermanet, Scott~E. Reed,
  Dragomir Anguelov, Dumitru Erhan, Vincent Vanhoucke, and Andrew Rabinovich.
\newblock Going deeper with convolutions.
\newblock {\em 2015 IEEE Conference on Computer Vision and Pattern Recognition
  (CVPR)}, pages 1--9, 2015.

\bibitem{Szegedy2013IntriguingPO}
Christian Szegedy, Wojciech Zaremba, Ilya Sutskever, Joan Bruna, Dumitru Erhan,
  Ian~J. Goodfellow, and Rob Fergus.
\newblock Intriguing properties of neural networks.
\newblock {\em CoRR}, abs/1312.6199, 2013.

\bibitem{Xie_2019_CVPR}
Cihang Xie, Zhishuai Zhang, Yuyin Zhou, Song Bai, Jianyu Wang, Zhou Ren, and
  Alan~L. Yuille.
\newblock Improving transferability of adversarial examples with input
  diversity.
\newblock In {\em The IEEE Conference on Computer Vision and Pattern
  Recognition (CVPR)}, June 2019.

\bibitem{Yuan2017AdversarialEA}
Xiaoyong Yuan, Pan He, Qile Zhu, Rajendra~Rana Bhat, and Xiaolin Li.
\newblock Adversarial examples: Attacks and defenses for deep learning.
\newblock {\em CoRR}, abs/1712.07107, 2017.

\bibitem{Zhou2018TransferableAP}
Wen Zhou, Xin Hou, Yongjun Chen, Mengyun Tang, Xiangqi Huang, Xiang Gan, and
  Yong Yang.
\newblock Transferable adversarial perturbations.
\newblock In {\em ECCV}, 2018.

\end{thebibliography}
}

\clearpage
\newpage 

\section*{Appendix}
\appendix

This supplementary file consists of:
\begin{itemize}
    \item A more thorough description of the networks used, the layers selected for attack, and full results on other attacks tested
    \item Complete disturbance graphs across layers of an attacking comprised of ILAP and I-FGSM
    \item A more complete result showing the comparison between ILAP and ILAF
    \item Full results for ILAP's performance on ImageNet
    \item Visualization of decision boundary
    \item Results for different $L_\infty$ norm values
    \item Results for ablating the learning rate used in ILAP
    \item Results comparing ILAP to TAP \cite{Zhou2018TransferableAP} on CIFAR-10
\end{itemize}

\section{ILAP Network Overview and Results for Other Base Attacks}
\label{ILAP_cifar10}
As shown in the main paper, we tested ILAP against MI-FGSM, C\&W, and TAP. We also tested I-FGSM, DeepFool, and FGSM. We test on a variety of models, namely: ResNet18 \cite{He2016DeepRL}, SENet18 \cite{Hu2017SqueezeandExcitationN}, DenseNet121\cite{Huang2017DenselyCC} and GoogLeNet \cite{Szegedy2015GoingDW} trained on CIFAR-10. For each source model, each large block output in the source model and each attack $A$, we generate adversarial examples for all images in the test set using $A$ with 20 iterations as a baseline. We then generate adversarial examples using $A$ with 10 iterations as input to ILA, which will then run for 10 iterations. The learning rate is set to $0.002$ for I-FGSM, $0.002$ for I-FGSM with momentum and $0.006$ for ILAP. We are in the $L_\infty$ norm setting with $\epsilon = 0.015$ for all attacks. We then evaluate transferability of baseline and ILA adversarial examples over the other models by testing their accuracies, as shown in Figure \ref{fig:ILAP-ciar10}.

Below is the list of layers (models from \cite{Liu2018}) we picked for each source model, which is indexed starting from 0 in the experiment results: 

\begin{itemize}
    \item ResNet18: conv, bn, layer1, layer2, layer3, layer4, linear (layer1-4 are basic blocks)
    \item GoogLeNet: pre\_layers, a3, b3, maxpool, a4, b4, c4, d4, e4, a5, b5, avgpool, linear
    \item DenseNet121: conv1, dense1, trans1, dense2, trans2, dense3, trans3, dense4, bn, linear
    \item SENet18: conv1, bn1, layer1, layer2, layer3, layer4, linear (layer1-4 are pre-activation blocks)
\end{itemize}

Additional results for the I-FGSM, FGSM, and DeepFool attacks are given in tables \ref{tab:IFGSM} and \ref{tab:FGSM-result}. Note that the output of DeepFool is clipped to satisfy our $\epsilon$-ball constraint. 

\begin{table*}[!htb]
  \caption{\label{tab:IFGSM} ILAP vs. I-FGSM and DeepFool Results}
  \centering
  
  \begin{tabular}{cccccccc}
    \toprule
    & & \multicolumn{3}{c}{I-FGSM} & \multicolumn{3}{c}{DeepFool} \\
    \cmidrule(r){3-5} \cmidrule(r){6-8}
    Source & Transfer & 20 Itr & 10 Itr ILAP & Opt ILAP  & 50 Itr & 25 Itr ILAP &  Opt ILAP \\
    \midrule
 & ResNet18\footnote[2]{Model that is exactly the same model as the source model.} & 3.3\% & 7.6\% & \textbf{1.8\%} (5)&48.7\% &            12.9\% &             \textbf{5.4\%} (5)\\
    ResNet18 & SENet18 & 44.4\% & \textbf{27.5\%} & \textbf{27.5\%} (4) &87.4\% &            \textbf{43.7\%} &            \textbf{43.7\%} (4)\\
    ($l=4$) & DenseNet121 & 45.8\% & \textbf{27.7\%} & \textbf{27.7\%} (4) &89.1\% &            \textbf{43.8\%} &            \textbf{43.8\%} (4)\\
    & GoogLeNet & 58.6\% & \textbf{35.8\%} & \textbf{35.8\%} (4)&89.3\% &            \textbf{50.7\%} &            \textbf{50.7\%} (4)\\
    \midrule
    
    & ResNet18 & 36.8\% & \textbf{25.8\%} & \textbf{25.8\%} (4)&91.9\% &            40.3\% &            \textbf{39.9\%} (5)\\
    SENet18 & $\text{SENet18}^\dagger$ & 2.4\% & 7.9\% & \textbf{2.3\%} (6)&56.8\% &            11.4\% &             \textbf{5.1\%} (6)\\
    ($l=4$) & DenseNet121 & 38.0\% & \textbf{25.9\%} & \textbf{25.9\%} (4) &92.9\% &            41.3\% &            \textbf{41.1\%} (5)\\
    & GoogLeNet & 48.4\% & \textbf{33.7\%} & \textbf{33.7\%} (4)&92.3\% &            \textbf{48.7\%} &            \textbf{48.7\%} (4)\\
    \midrule
    
    & ResNet18 & 45.1\% & \textbf{26.7\%} & \textbf{26.7\%}(6) &81.6\% &            \textbf{30.1\%} &            \textbf{30.1\%} (6)\\
    DenseNet121 & SENet18 & 43.4\% & \textbf{26.1\%} & \textbf{26.1\%}(6) &81.5\% &            29.0\% &            \textbf{28.9\%} (7)\\
    ($l=6$) & $\text{DenseNet121}^\dagger$ & 2.6\% & 1.7\% & \textbf{0.8\%}(9) &34.9\% &             4.1\% &             \textbf{3.3\%} (9)\\
    & GoogLeNet & 47.3\% & \textbf{28.6\%} & \textbf{28.6\%}(6) &82.3\% &            \textbf{32.4\%} &            \textbf{32.4\%} (6)\\
    \midrule
    
    & ResNet18 & 55.9\%& 34.0\%& \textbf{32.7\%}(3) &92.3\% &            \textbf{44.0\%} &            \textbf{44.0\%} (9)\\
    GoogLeNet & SENet18 & 55.6\% &33.1\%& \textbf{31.8\%} (3)&92.1\% &            \textbf{42.9\%} &            \textbf{42.9\%} (9)\\
    ($l=9$) & DenseNet121 & 48.9\% & 28.7\% & \textbf{28.1\%}(3)&93.1\% &            \textbf{38.1\%} &            \textbf{38.1\%} (9)\\
    & $\text{GoogLeNet}^\dagger$ & 0.9\% & 0.8\% & \textbf{0.4\%} (11)&51.5\% &             4.2\% &            \textbf{3.9\%} (11)\\

    \bottomrule
  \end{tabular}
\caption*{Table \ref{tab:IFGSM}. Accuracies after attack using ILAP based on I-FGSM and DeepFool. Note that although significant improvement for transfer is exhibited for DeepFool, the original attack transfer rates are quite poor (the accuracies are still quite high after a DeepFool transfer attack).}
\end{table*}

\begin{table*}[!htb]
  \caption{\label{tab:FGSM-result} ILAP vs. FGSM Results}
  \centering
  \begin{threeparttable}
\begin{tabular}{cccccccc}
\toprule
  & & \multicolumn{3}{c}{FGSM}  \\
    \cmidrule(r){3-5}\
    Source & Transfer & 20 Itr & 10 Itr ILAP & Opt ILAP  \\
\midrule
 & $\text{ResNet18}^\dagger$  &                                    47.7\% &             \textbf{2.0\%} &             \textbf{2.0\%} (6) \\
  ResNet18 & SENet18 &                                             63.6\% &            \textbf{42.6\%} &            \textbf{42.6\%} (6) \\
   ($l=6$) & DenseNet121 &                                         64.9\% &            44.6\% &            \textbf{44.5\%} (5) \\
   & GoogLeNet &                                                   66.5\% &            55.5\% &            \textbf{54.2\%} (4) \\

   \midrule
    &     ResNet18 &                                               60.7\% &            37.4\% &            \textbf{36.1\%} (5) \\
   SENet18 &      $\text{SENet18}^\dagger$ &                       40.7\% &             \textbf{3.0\%} &             \textbf{3.0\%} (6) \\
    ($l=6$) &  DenseNet121 &                                       61.8\% &            37.0\% &            \textbf{36.3\%} (5) \\
    &    GoogLeNet &                                               63.8\% &            46.3\% &            \textbf{45.3\%} (5) \\

   \midrule
 &     ResNet18 &                                                  65.0\% &            36.4\% &            \textbf{36.2\%} (6) \\
DenseNet121 &      SENet18 &                                       65.0\% &            \textbf{35.5\%} &            \textbf{35.5\%} (7) \\
 ($l=7$)  &  $\text{DenseNet121}^\dagger$ &                        47.3\% &             5.8\% &             \textbf{0.9\%} (9) \\
 &    GoogLeNet &                                                  64.6\% &            37.6\% &            \textbf{37.4\%} (6) \\

   \midrule
    &     ResNet18 &                                               64.9\% &            \textbf{43.5\%} &            \textbf{43.5\%} (9) \\
  GoogLeNet &      SENet18 &                                       65.1\% &            \textbf{43.8\%} &            \textbf{43.8\%} (9) \\
   ($l=9$)  &  DenseNet121 &                                       63.7\% &            \textbf{39.7\%} &            \textbf{39.7\%} (9) \\
  &    $\text{GoogLeNet}^\dagger$ &                                36.6\% &             5.9\% &            \textbf{0.6\%} (12) \\

\bottomrule
\end{tabular}
\begin{tablenotes}
  \item[$\dagger$] Same model as source model.
  \end{tablenotes}
\end{threeparttable}
\caption*{Table \ref{tab:FGSM-result}. Accuracies after attack based on FGSM. Note that significant improvement occurs in the ILAP settings.}
\end{table*}

\begin{figure*}[!htb] 
  \centering
  \includegraphics[width=0.9\textwidth]{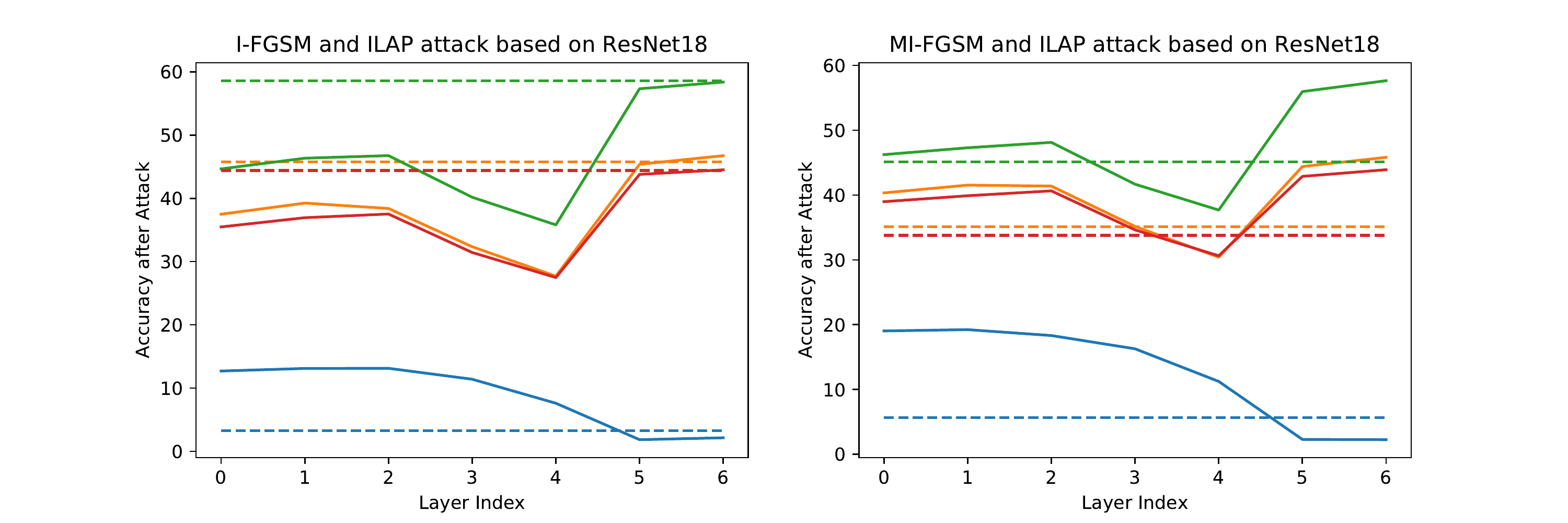} 
  \includegraphics[width=0.9\textwidth]{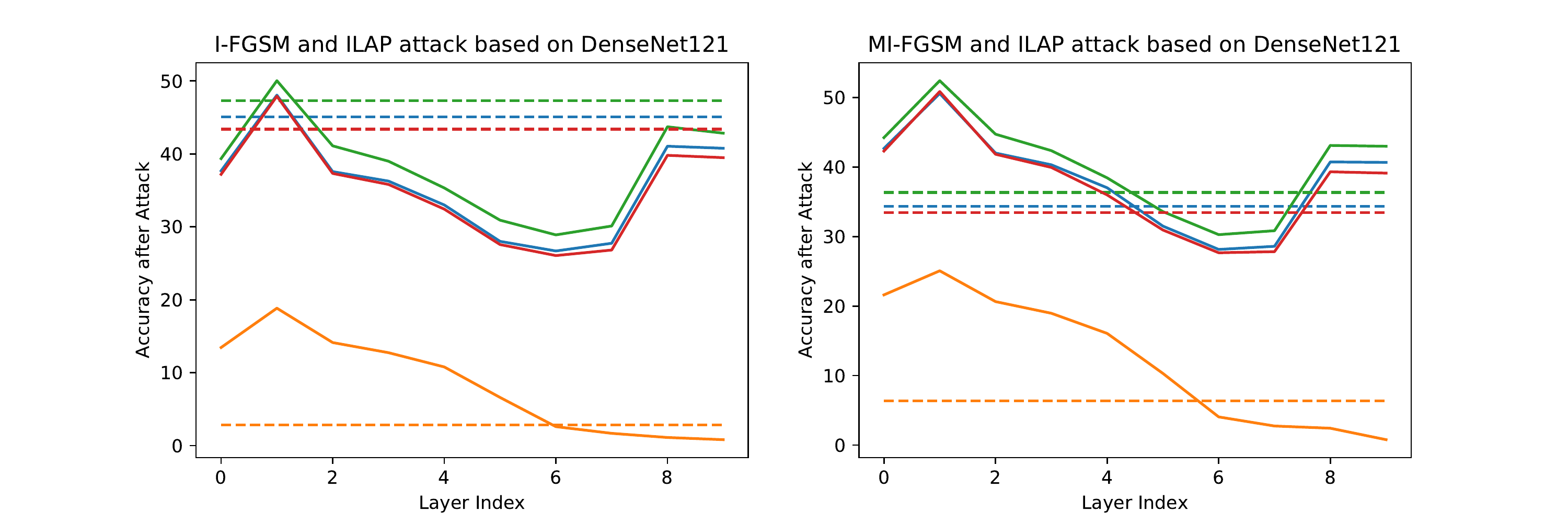}
  \includegraphics[width=0.9\textwidth]{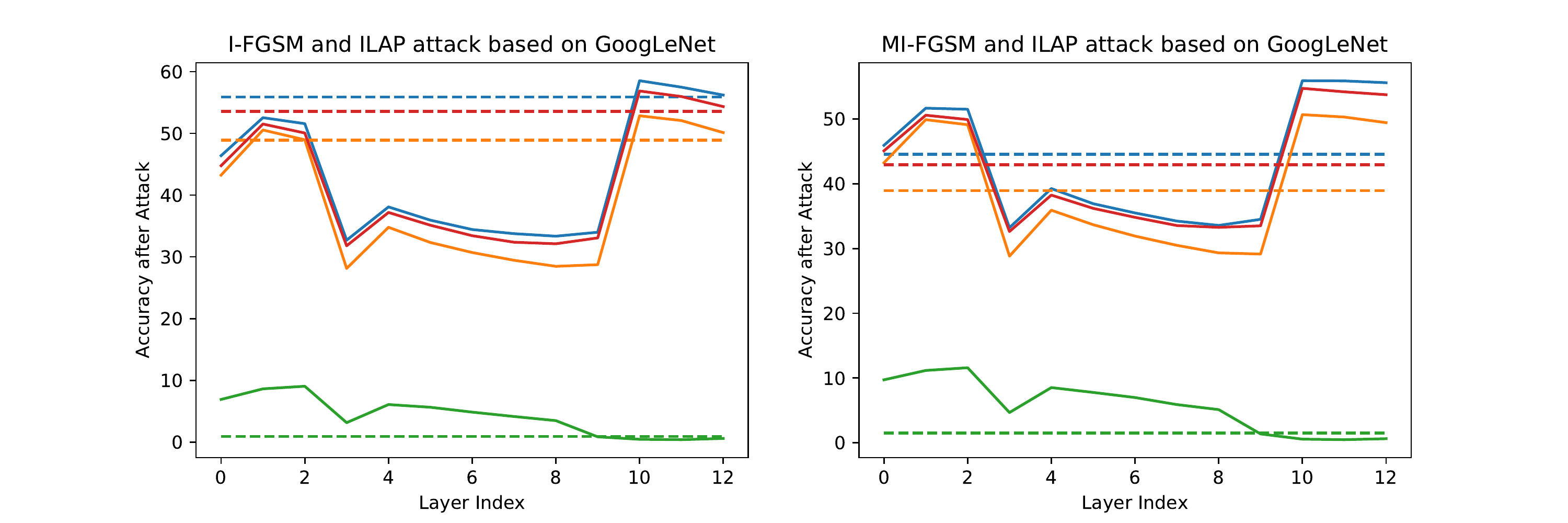}
  \includegraphics[width=0.9\textwidth]{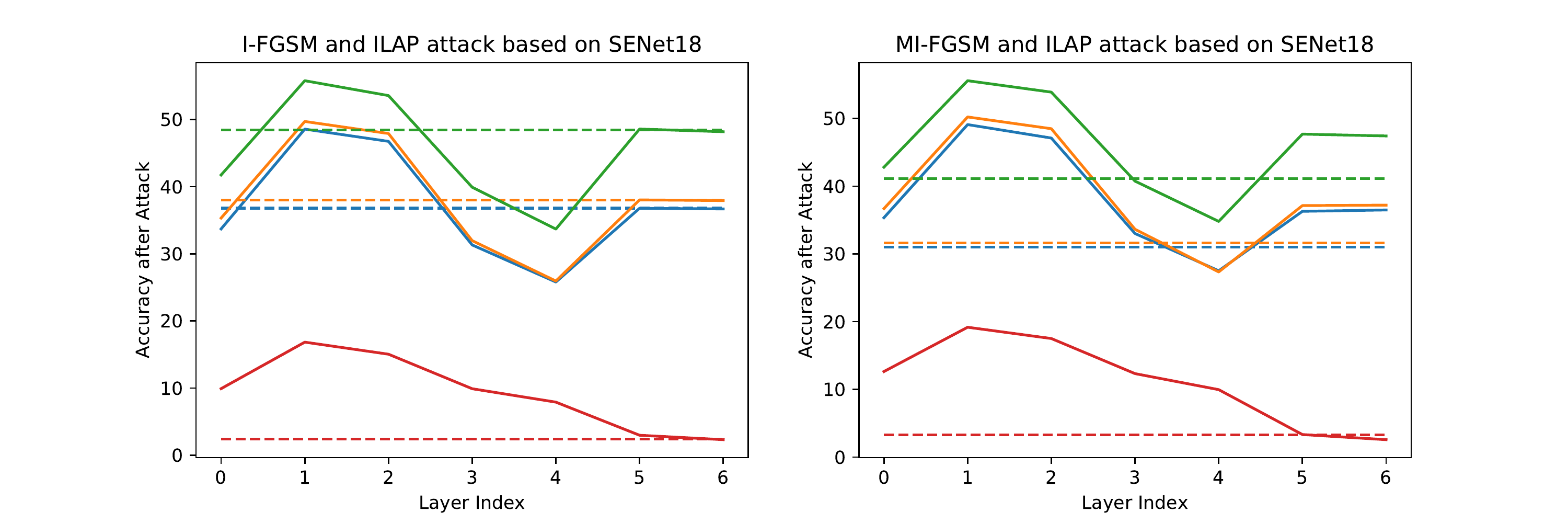}
  \includegraphics[width=0.2\textwidth]{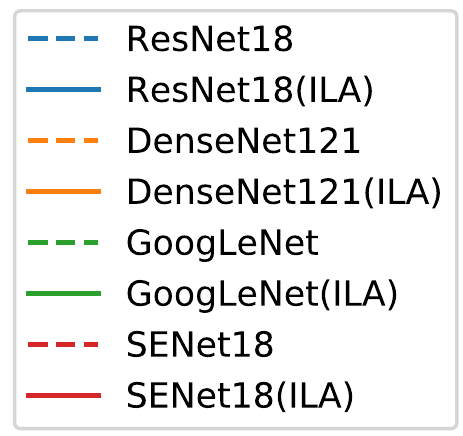}
  \caption{\label{fig:ILAP-ciar10}Visualizations for ILAP against I-FGSM and MI-FGSM baselines on CIFAR-10}
\end{figure*}

\begin{figure*}[!htb] 
  \centering
  \includegraphics[width=0.9\textwidth]{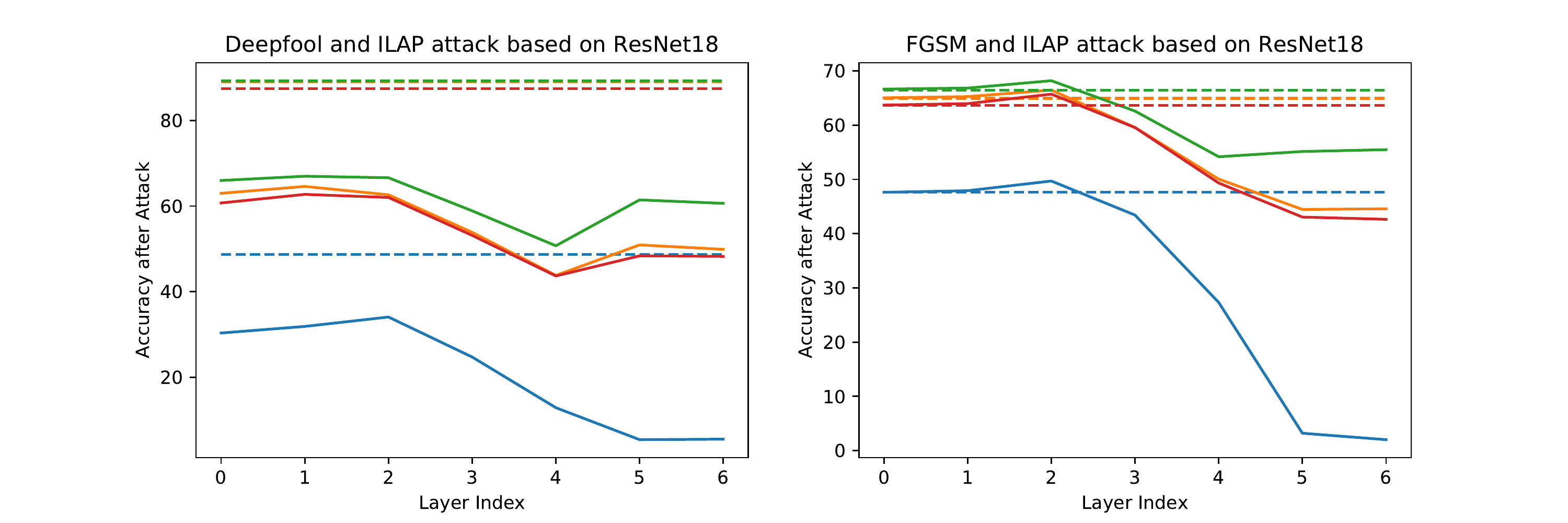}
  \includegraphics[width=0.9\textwidth]{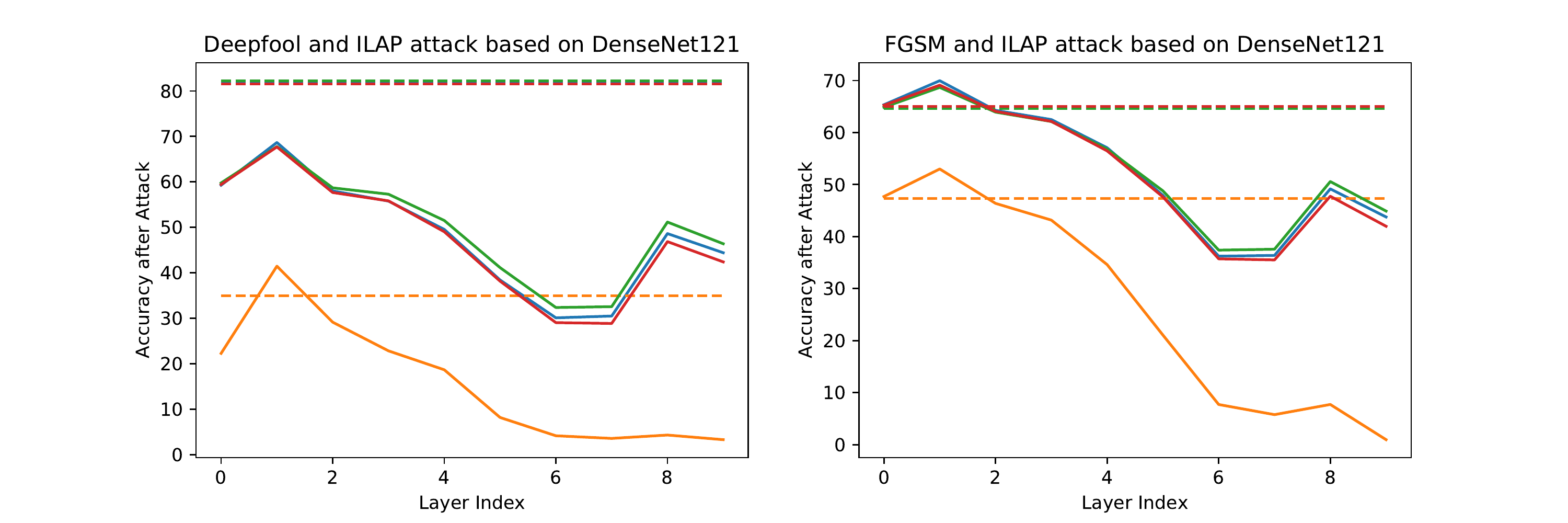}
  \includegraphics[width=0.9\textwidth]{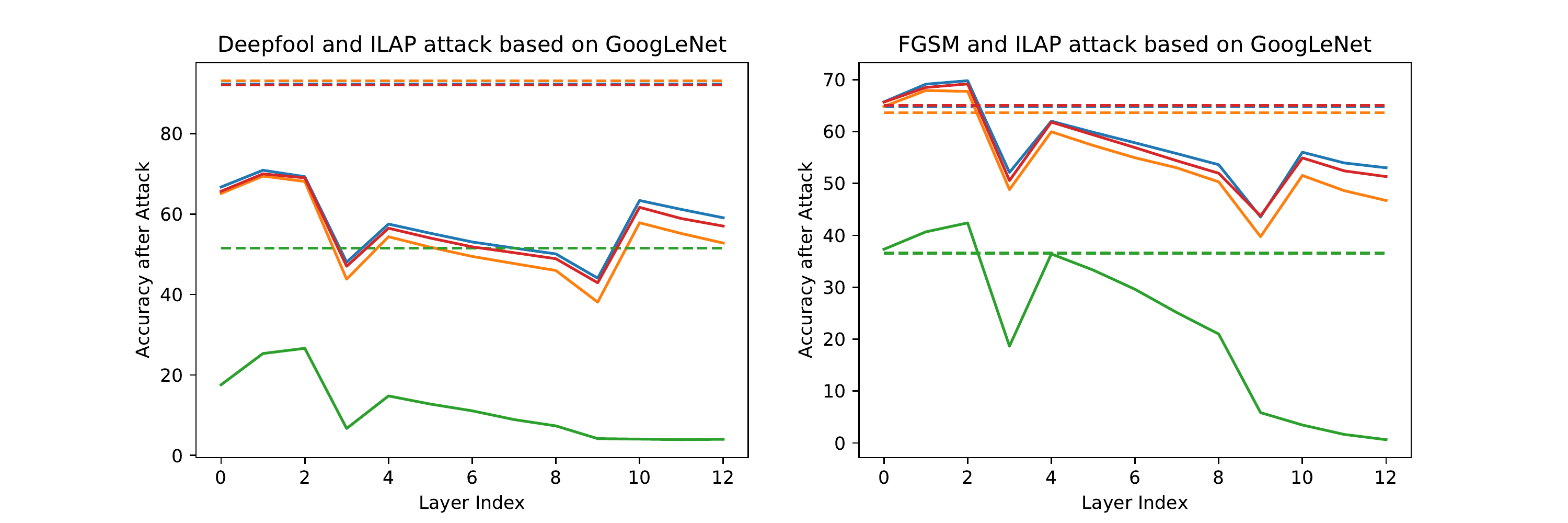}
  \includegraphics[width=0.9\textwidth]{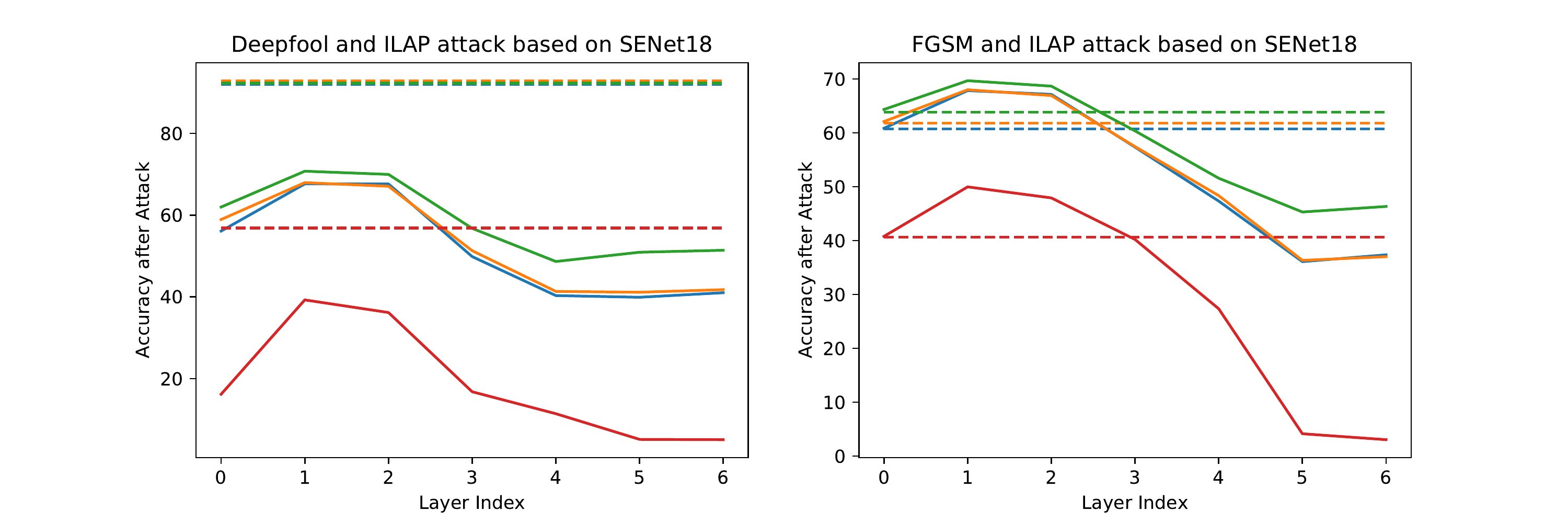}
  \includegraphics[width=0.2\textwidth]{appdx_figures/transferability_legend.pdf}
  \caption{\label{fig:ILAP-ciar10}Visualizations for ILAP aganist Deepfool and FGSM with momentum baselines on CIFAR-10}
\end{figure*}

\begin{figure*}[!htb] 
  \centering
  \includegraphics[width=0.9\textwidth]{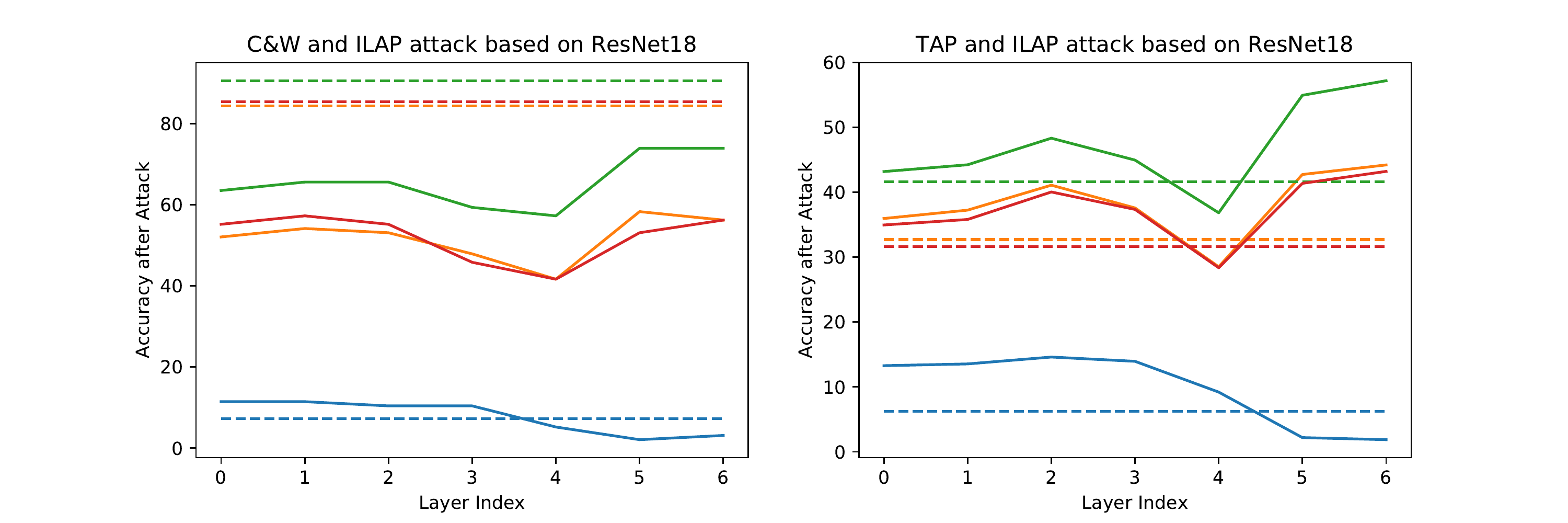}
  \includegraphics[width=0.9\textwidth]{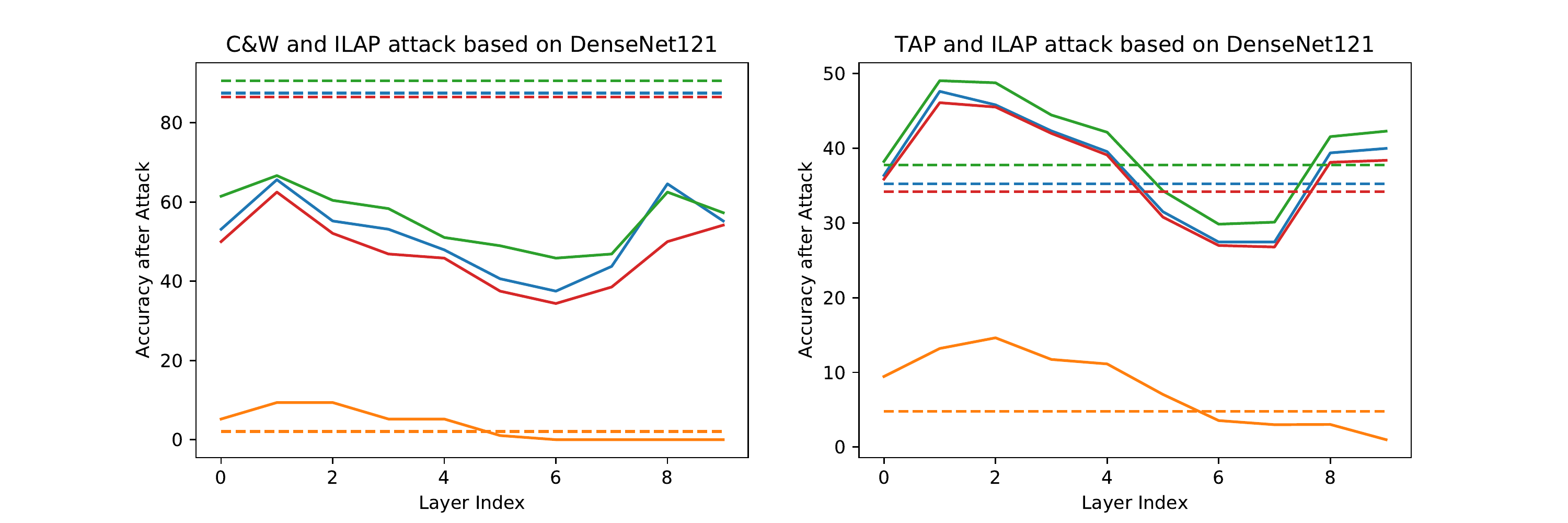}
  \includegraphics[width=0.9\textwidth]{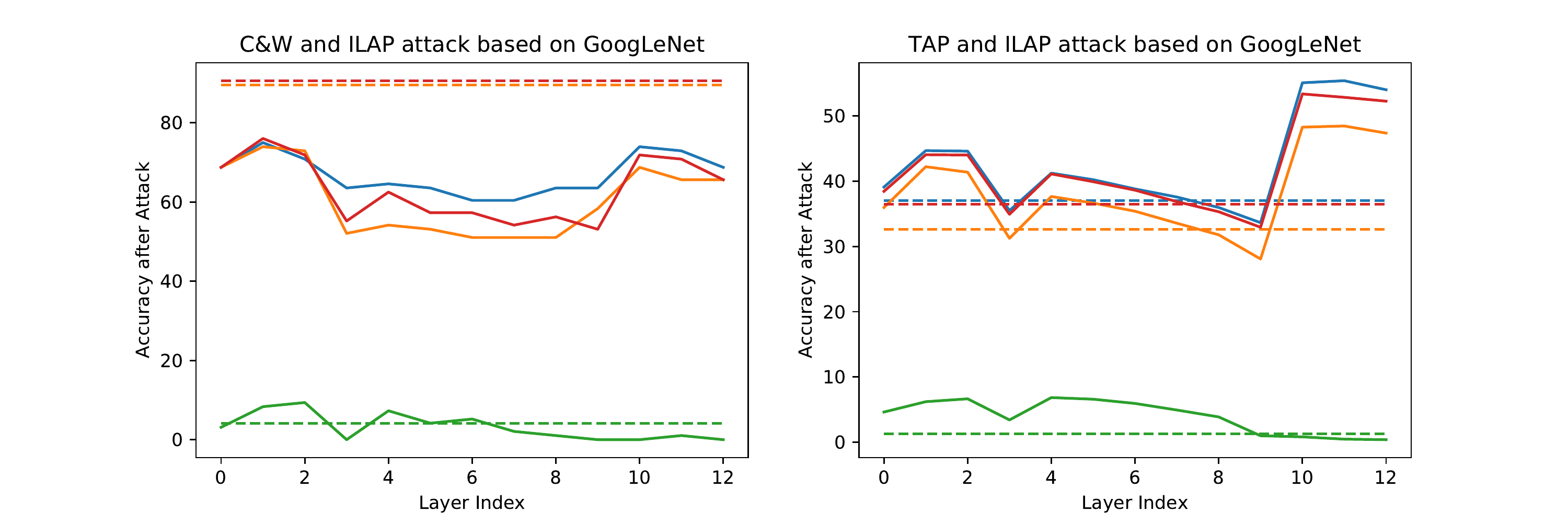}
  \includegraphics[width=0.9\textwidth]{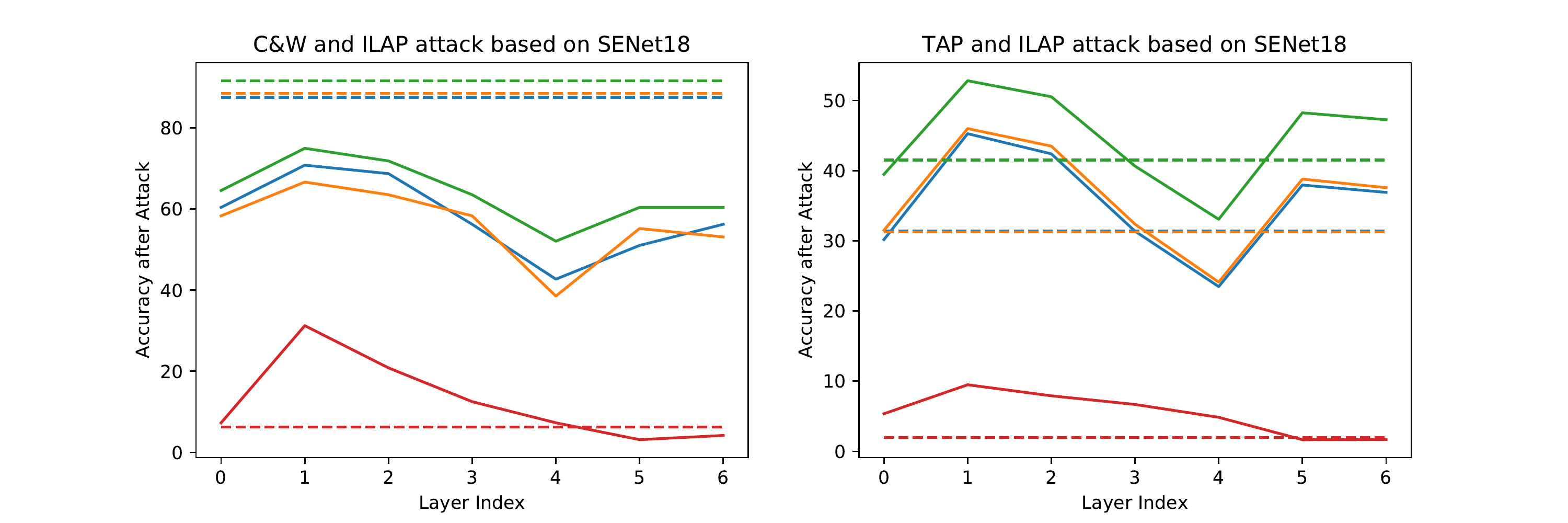}
  \includegraphics[width=0.2\textwidth]{appdx_figures/transferability_legend.pdf}
  \caption{\label{fig:ILAP-ciar10}Visualizations for ILAP aganist Deepfool and FGSM with momentum baselines on CIFAR-10}
\end{figure*}

\section{Disturbance graphs}

In this experiment, we used the same setting as our main experiment in Appendix \ref{ILAP_cifar10} to generate adversarial examples, with only I-FGSM used as the reference attack. The average disturbance of each set of adversarial examples is calculated at each layer. We repeated the experiment for all four models described in Appendix \ref{ILAP_cifar10}, as shown in Figure \ref{fig:dist}. Observe that the $l$ in the legend refers to the hyperparameter set in the ILA attack, and afterwards the disturbance values were computed on layers indicated by the $l$ in the x-axis.

\begin{figure}[!htb]
  \centering
  \includegraphics[width=0.5\textwidth]{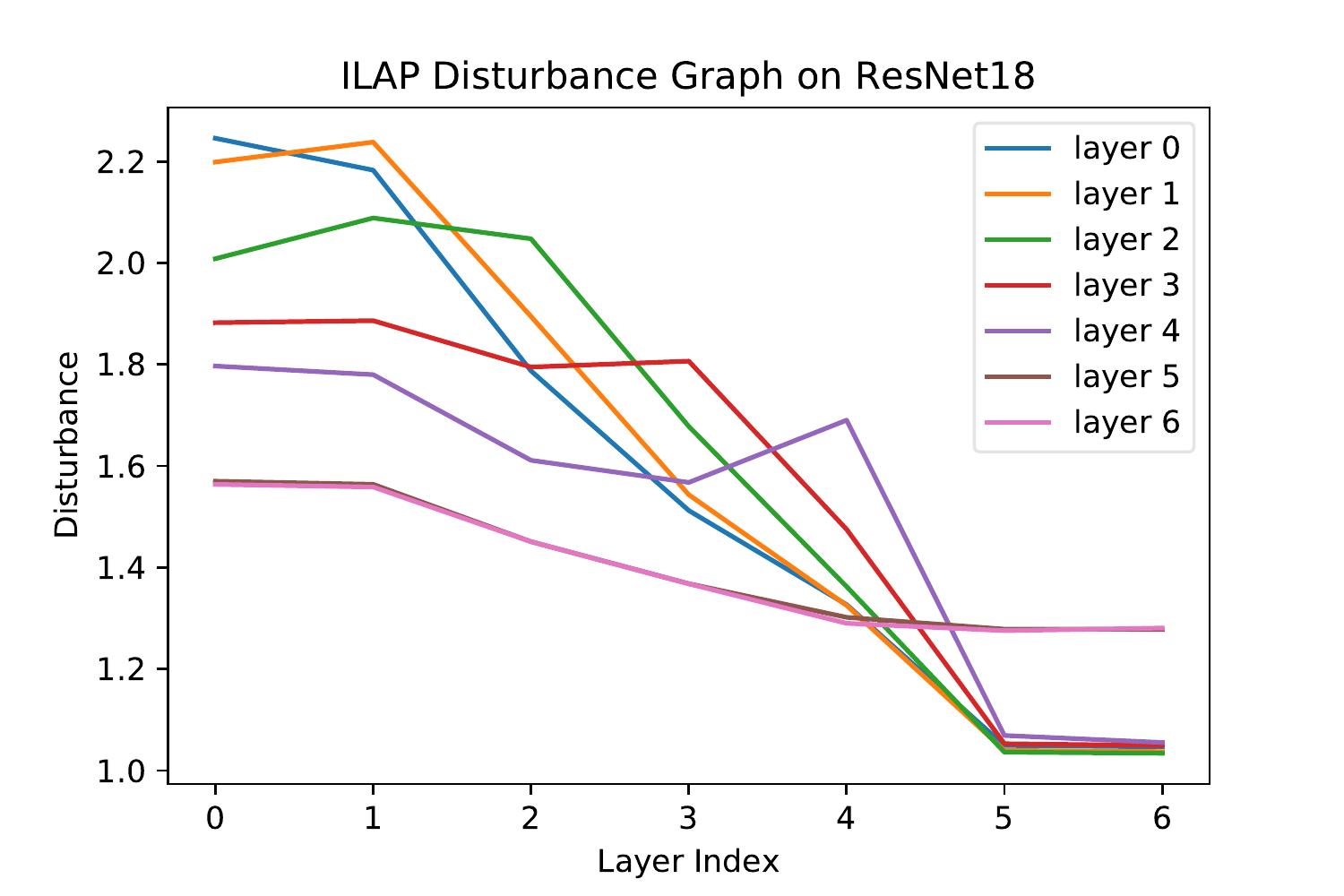}
  \includegraphics[width=0.5\textwidth]{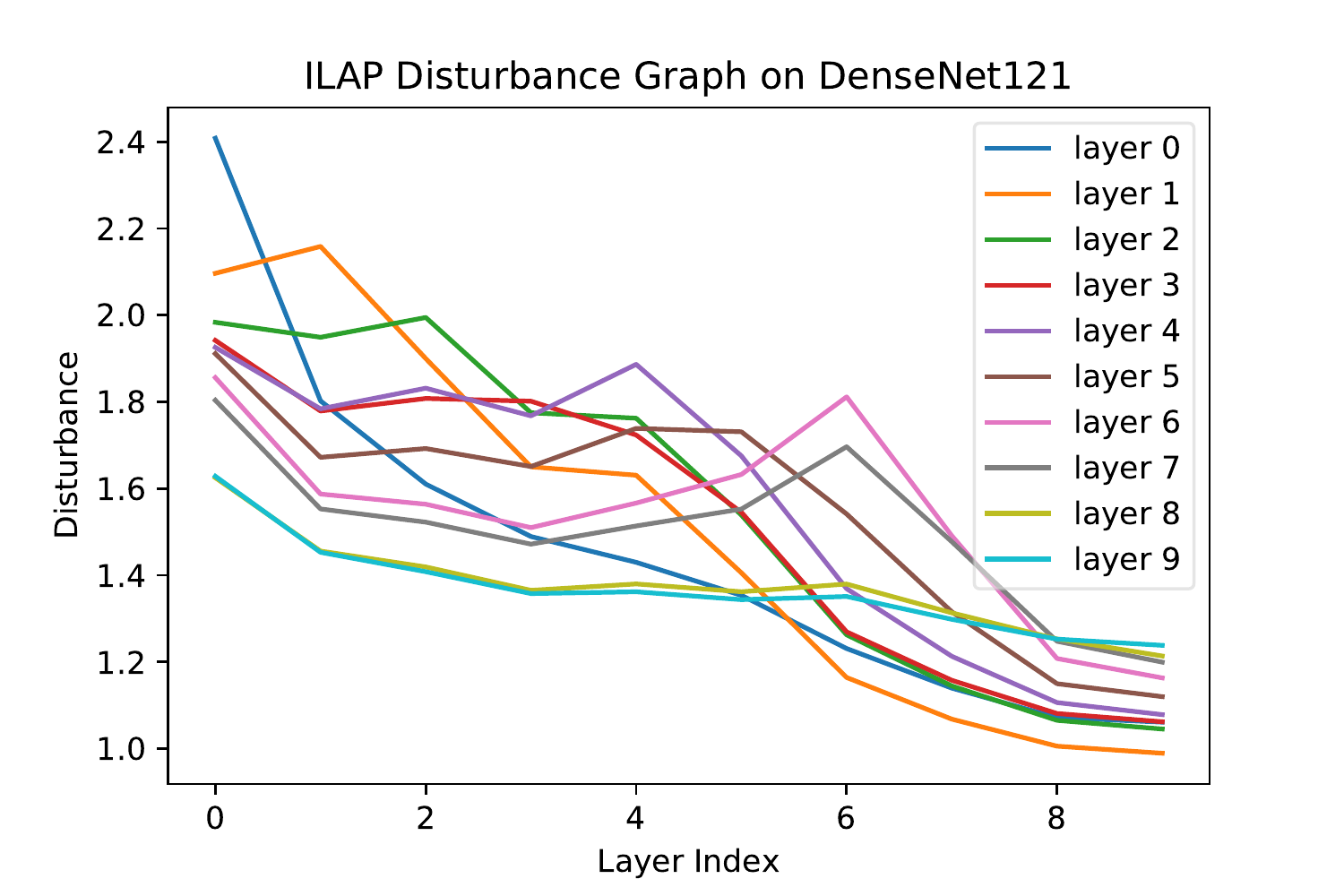}
  \includegraphics[width=0.5\textwidth]{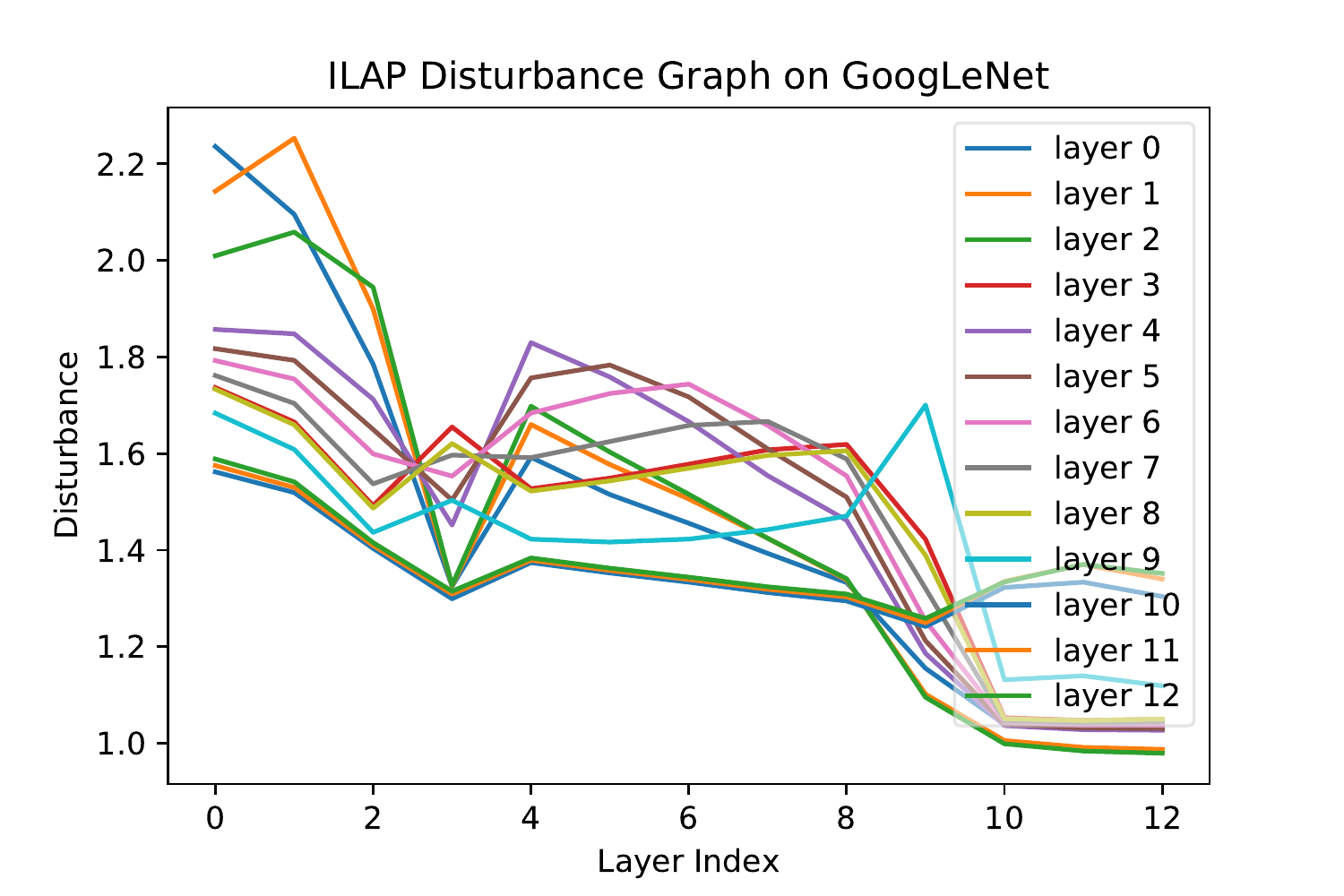}
  \includegraphics[width=0.5\textwidth]{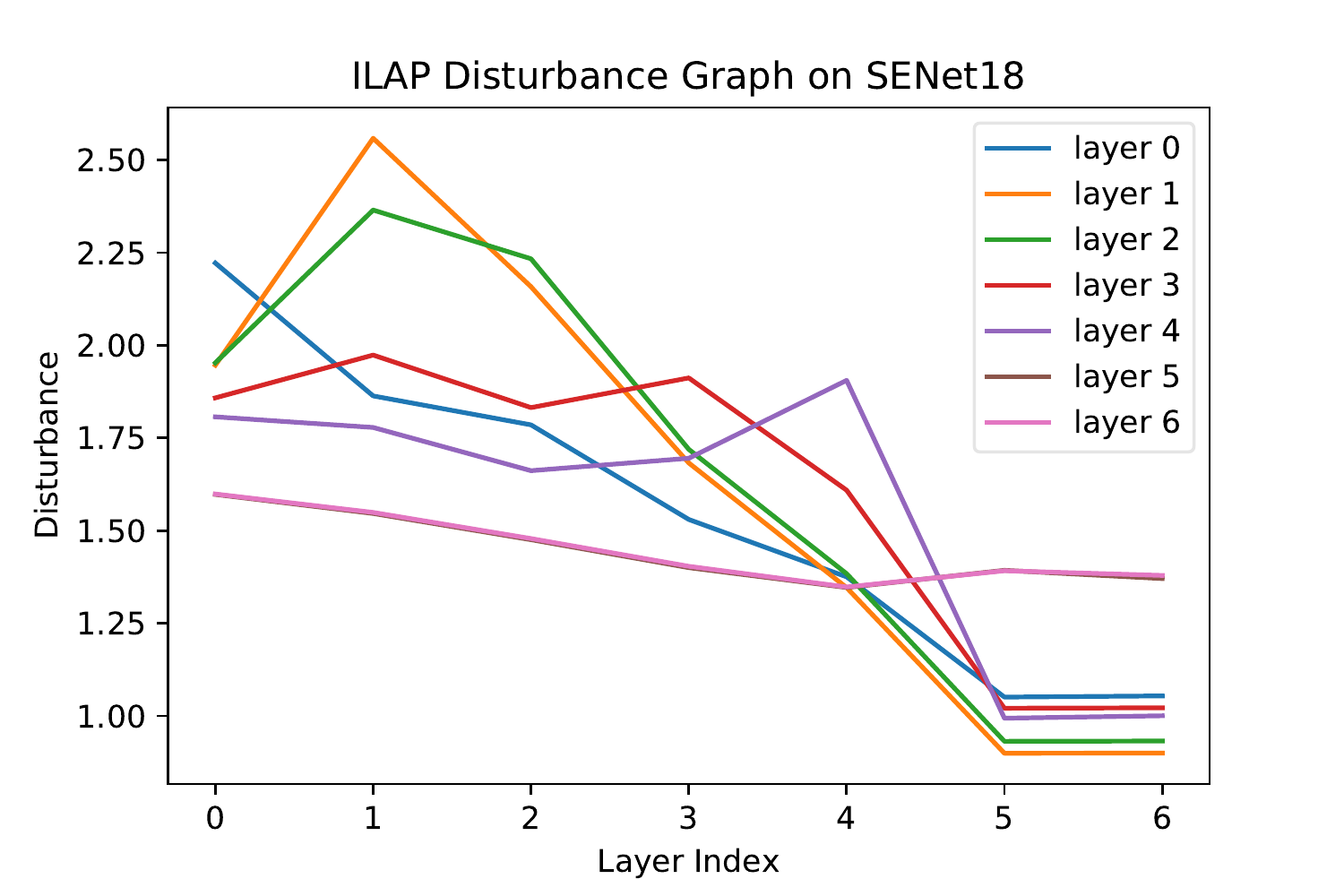}
  \caption{\label{fig:dist} Disturbance graphs of ILAP with I-FGSM as reference}
\end{figure}

\section{ILAP vs ILAF Full Result}

\begin{figure}[!htb]
  \centering
  \includegraphics[width=0.45\textwidth]{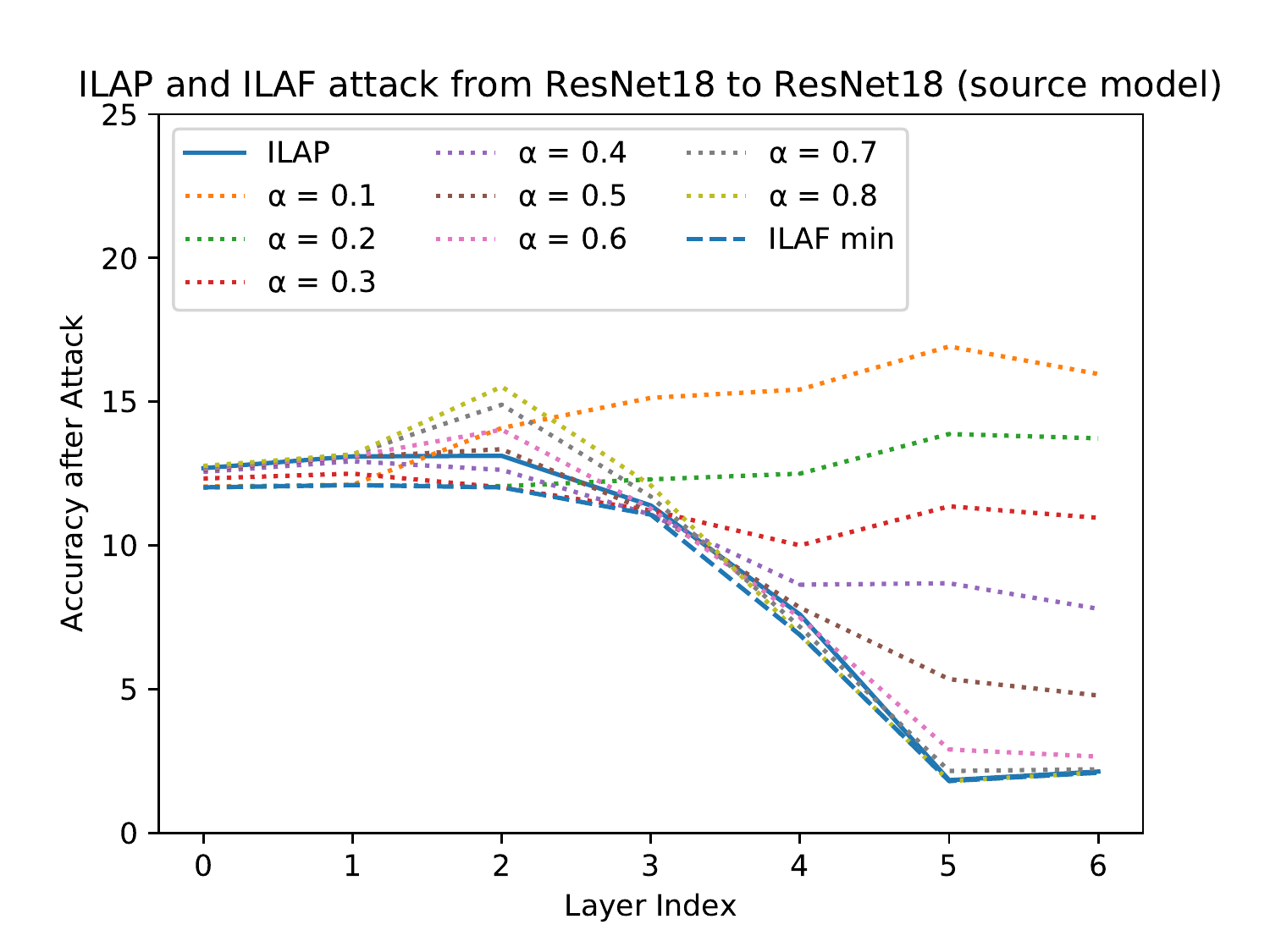}
  \includegraphics[width=0.45\textwidth]{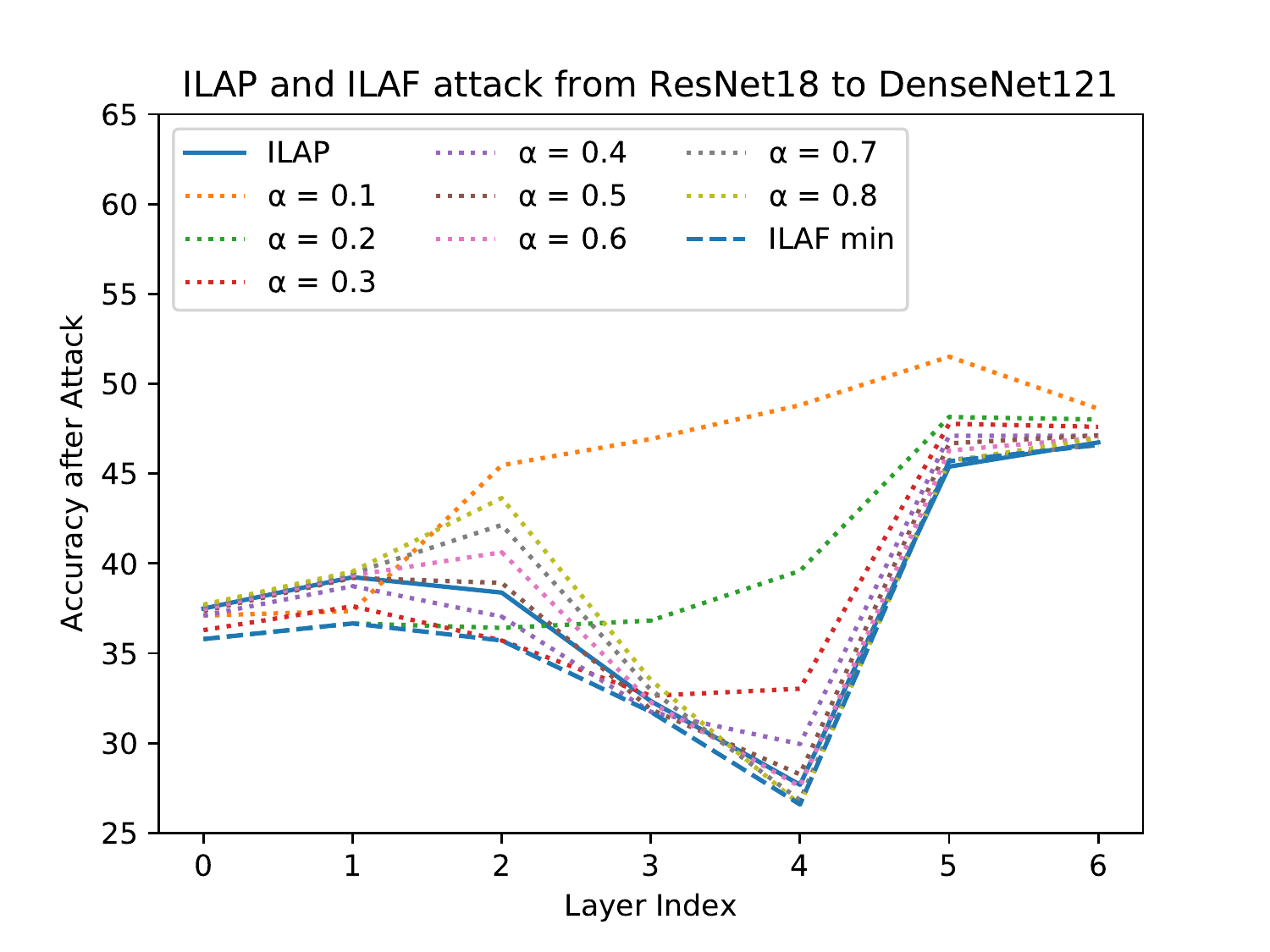}
  \includegraphics[width=0.45\textwidth]{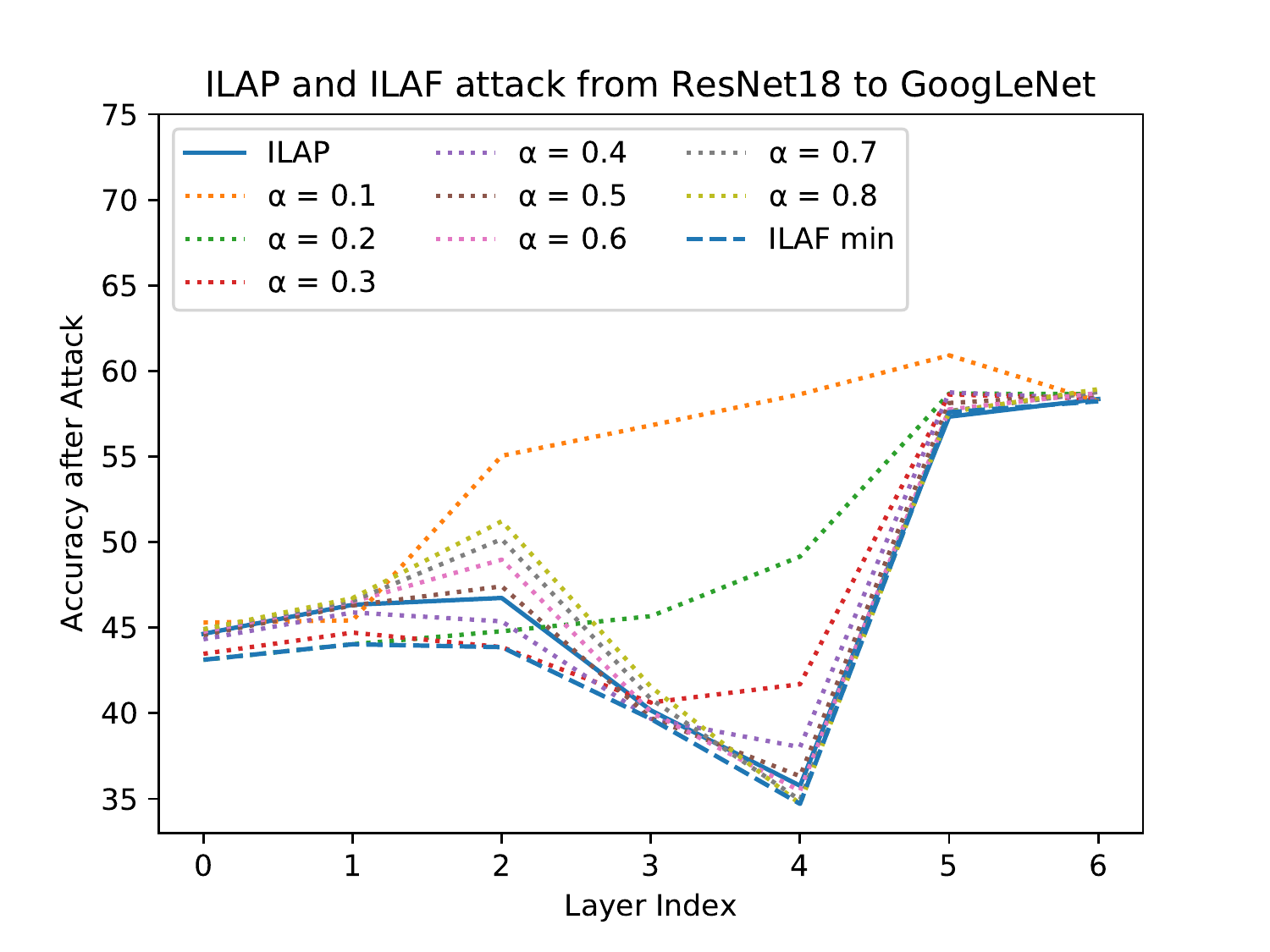}
  \includegraphics[width=0.45\textwidth]{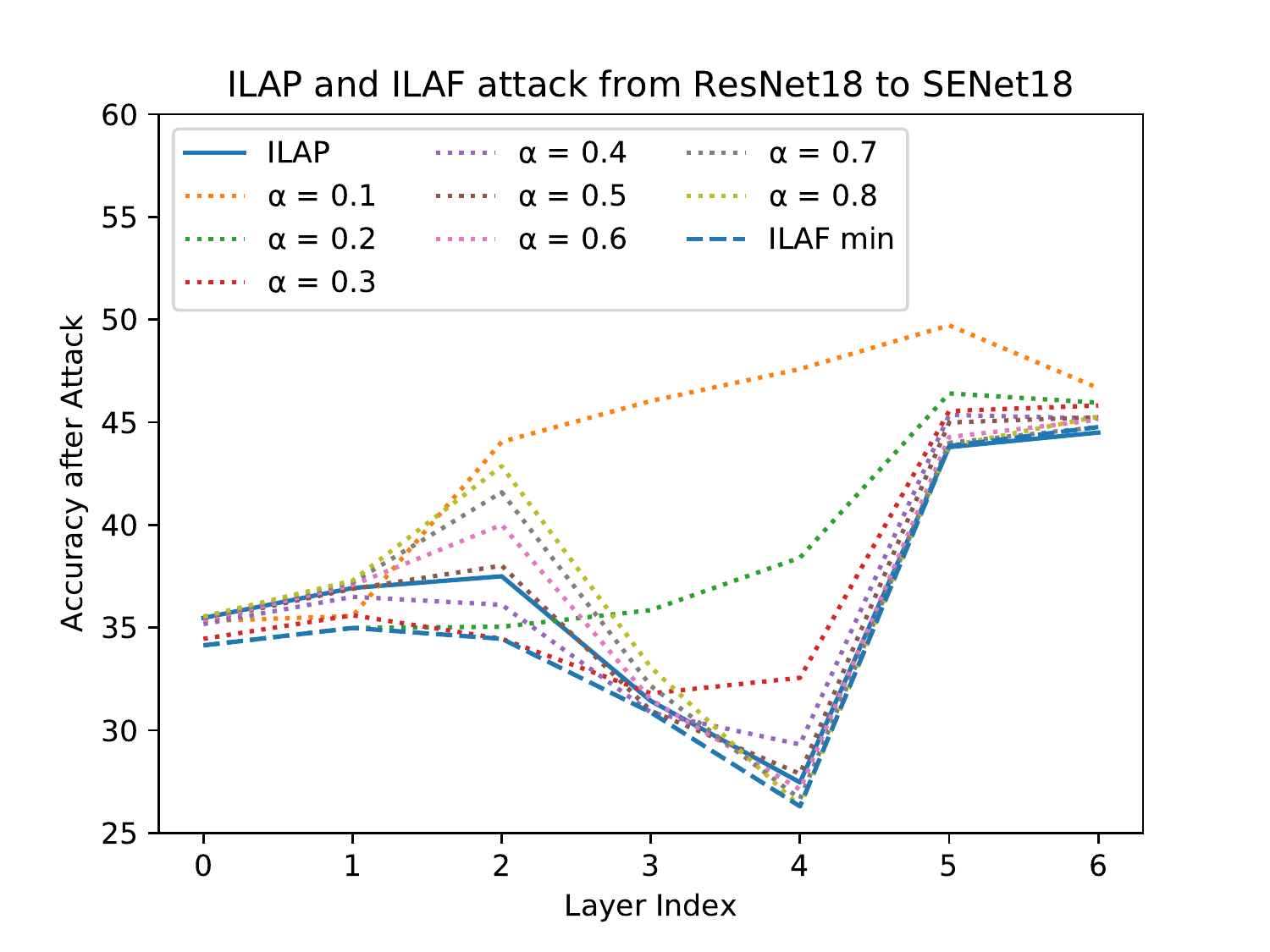}
  \caption{\label{fig:compare} ILAP vs ILAF comparisions}
\end{figure}

As described in the main paper, we compared the performace of ILAP and ILAF with a range of $\alpha$. We used the same setting as our main experiment in Appendix \ref{ILAP_cifar10} for ILAP and ILAF to generate adversarial examples, with only I-FGSM used as the reference attack. The result is shown in Figure \ref{fig:compare}.  

\section{ILAP on ImageNet Full Result}

We tested ILAP against I-FGSM and I-FGSM with momentum on ImageNet similarly to the experiment on CIFAR-10. The models we used are ResNet18, DenseNet121, SqueezeNet1.0 and AlexNet. The learning rate is set to $0.008$ for I-FGSM, $0.01$ for ILAP plus I-FGSM, $0.018$ for I-FGSM with momentum  and $0.018$ for ILAP plus I-FGSM with momentum. To evaluate transferability, we test the accuracies of different models over adversarial examples generated from all $50000$ ImageNet test images, as shown in Figure \ref{fig:imagenet} \footnote{Previous versions of this paper included results with incorrect normalization.}.

Below is the list of layers (models from \cite{Marcel2010TorchvisionTM}) we picked for each source model: 

\begin{itemize}
    \item ResNet18: conv1, bn1, layer1, layer2, layer3, layer4, fc 
    
    \item DenseNet121: conv0, denseblock1, transition1, denseblock2, transition2, denseblock3, transition3, denseblock4, norm5, classifier
    \item SqueezeNet1.0: Features: 0 3 4 5 7 8 9 10 12, classifier 
    \item AlexNet: Features: 0 3 4 6 8 10, classifiers: 1 4
\end{itemize}

\begin{figure*}[!htb]
  \centering
  \includegraphics[width=0.9\textwidth]{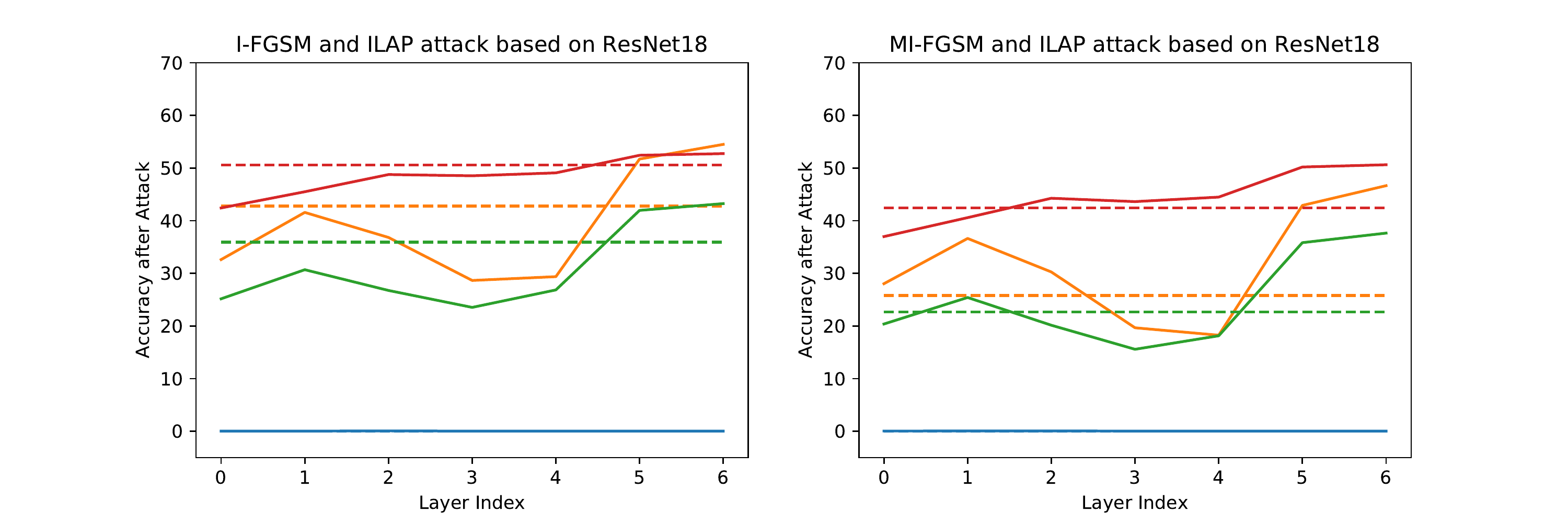}
  \includegraphics[width=0.9\textwidth]{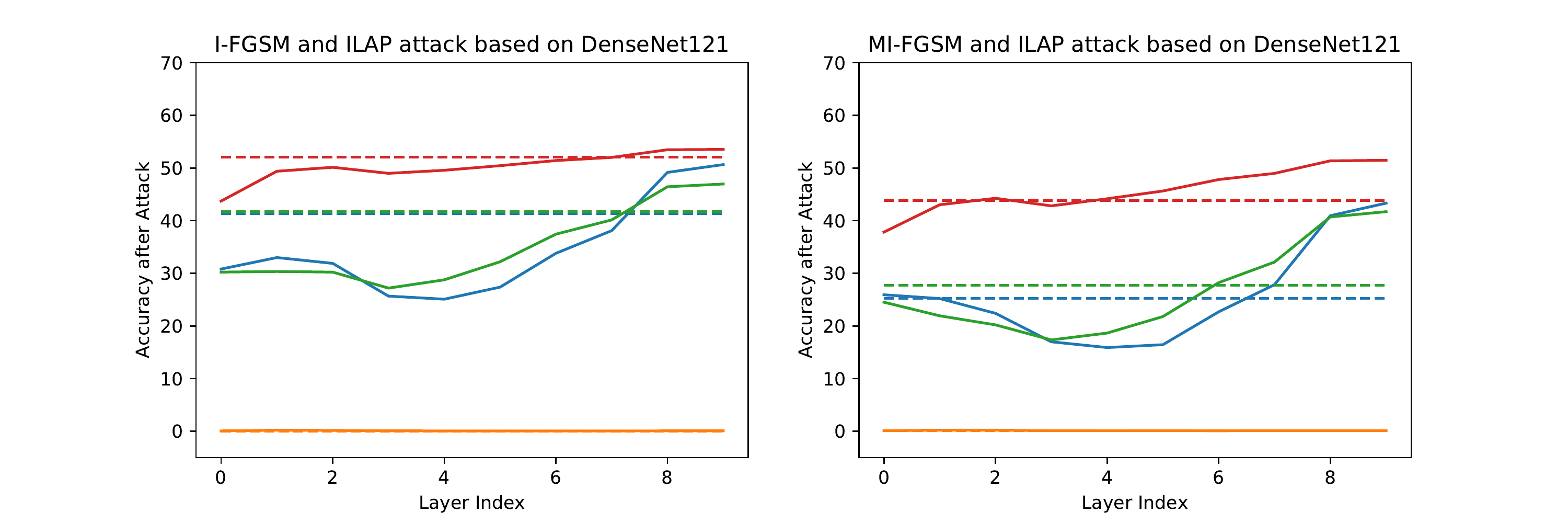}
  \includegraphics[width=0.9\textwidth]{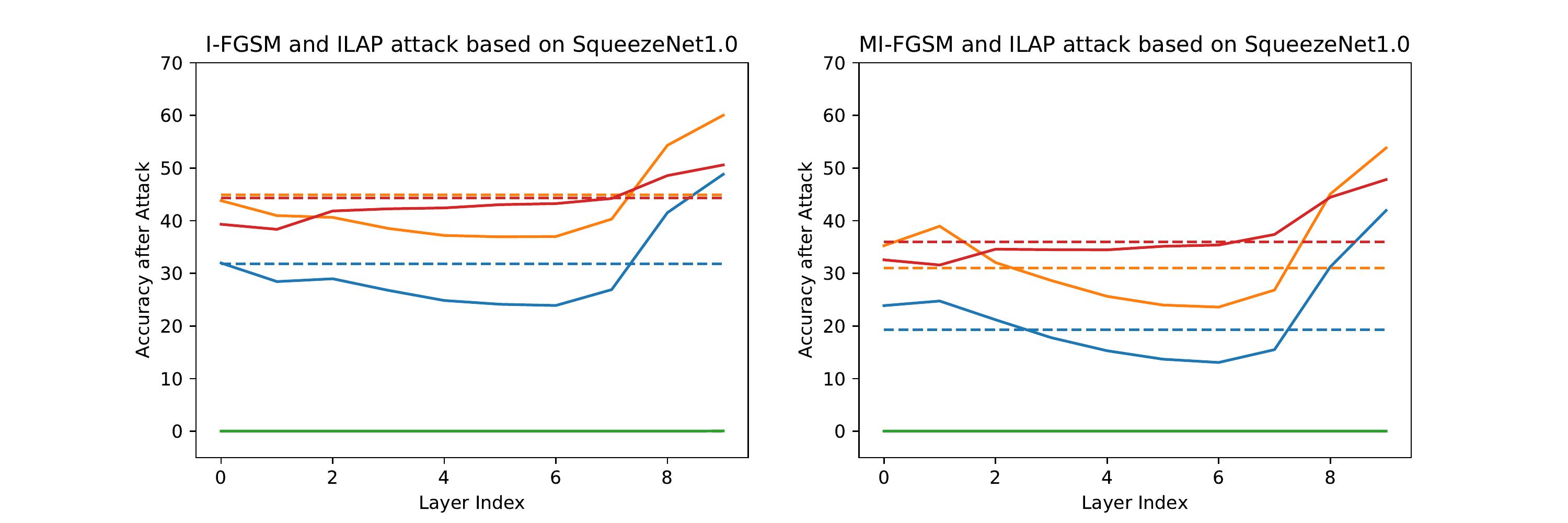}
  \includegraphics[width=0.9\textwidth]{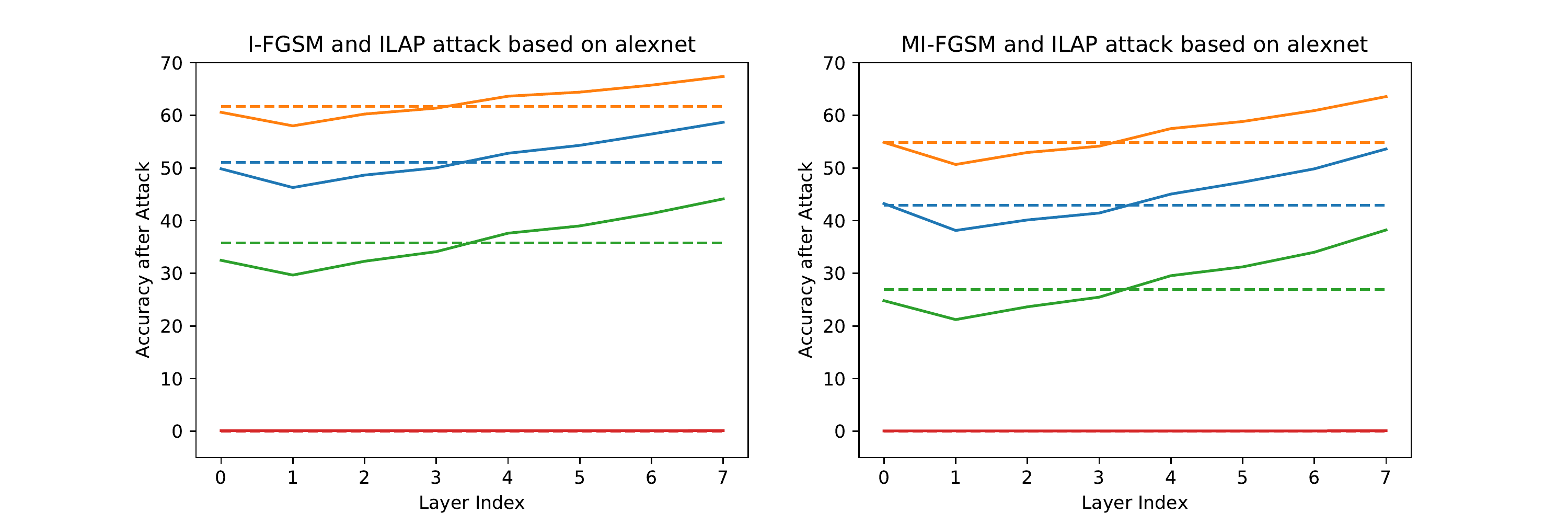}
  \includegraphics[width=0.2\textwidth]{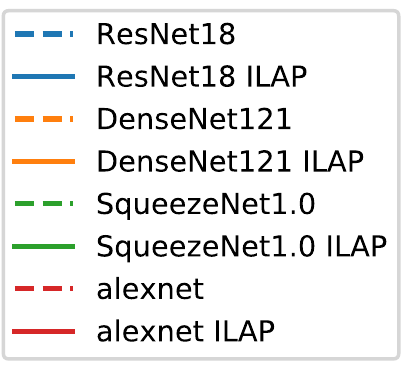}
  \caption{\label{fig:imagenet} Visualizations for ILAP against I-FGSM and I-FGSM with momentum baselines on ImageNet }
\end{figure*}   

\section{Visualization of the Decision Boundary}
\begin{figure}[!htb]
  \begin{minipage}[t]{0.40\textwidth}
    \centering
    \includegraphics[width=\textwidth]{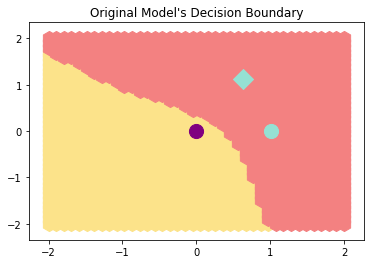}
  \end{minipage}
  \hspace{-0.1cm}
  \hfill
  \begin{minipage}[t]{0.40\textwidth}
    \centering
    \includegraphics[width=\textwidth]{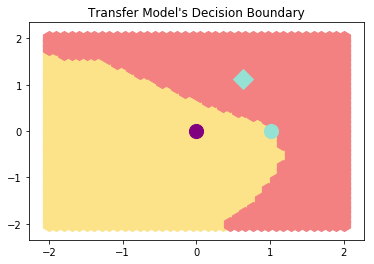}
  \end{minipage}
  \begin{minipage}[t]{0.40\textwidth}
    \centering
    \includegraphics[width=\textwidth]{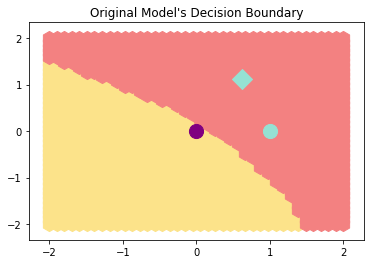}
  \end{minipage}
  \hspace{-0.1cm}
  \hfill
  \begin{minipage}[t]{0.40\textwidth}
    \centering
    \includegraphics[width=\textwidth]{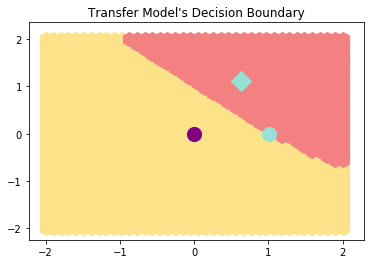}
  \end{minipage}
  \centering
  \caption{\label{fig:visualize} Visualization of the decision boundary relative to the two adversarial examples generated. Yellow is the correct label's decision space, red is the incorrect label's decision space. The purple dot is the original image's location, the green circle is the I-FGSM perturbation, and the green diamond is the ILA perturbation. Note that for the above, it seems the vector between the purple dot and green diamond is more orthogonal to the decision boundary than the vector between the purple dot and green dot (hence roughly indicating that ILA is working as intended in producing a more orthogonal transfer vector). }
\end{figure}

To gain some understanding over how ILA interplays with the decision boundaries, we visualize the two dimensional plane between the initial I-FGSM perturbation and the ILA perturbation for some examples. Visualization is done on Resnet with layer 4, and I-FGSM as the starting perturbation. See Figure \ref{fig:visualize}.

\section{Fooling with Different $L_\infty$ Values}
\begin{figure}[!htb]
  \begin{minipage}[t]{0.45\textwidth}
    \centering
    \includegraphics[width=\textwidth]{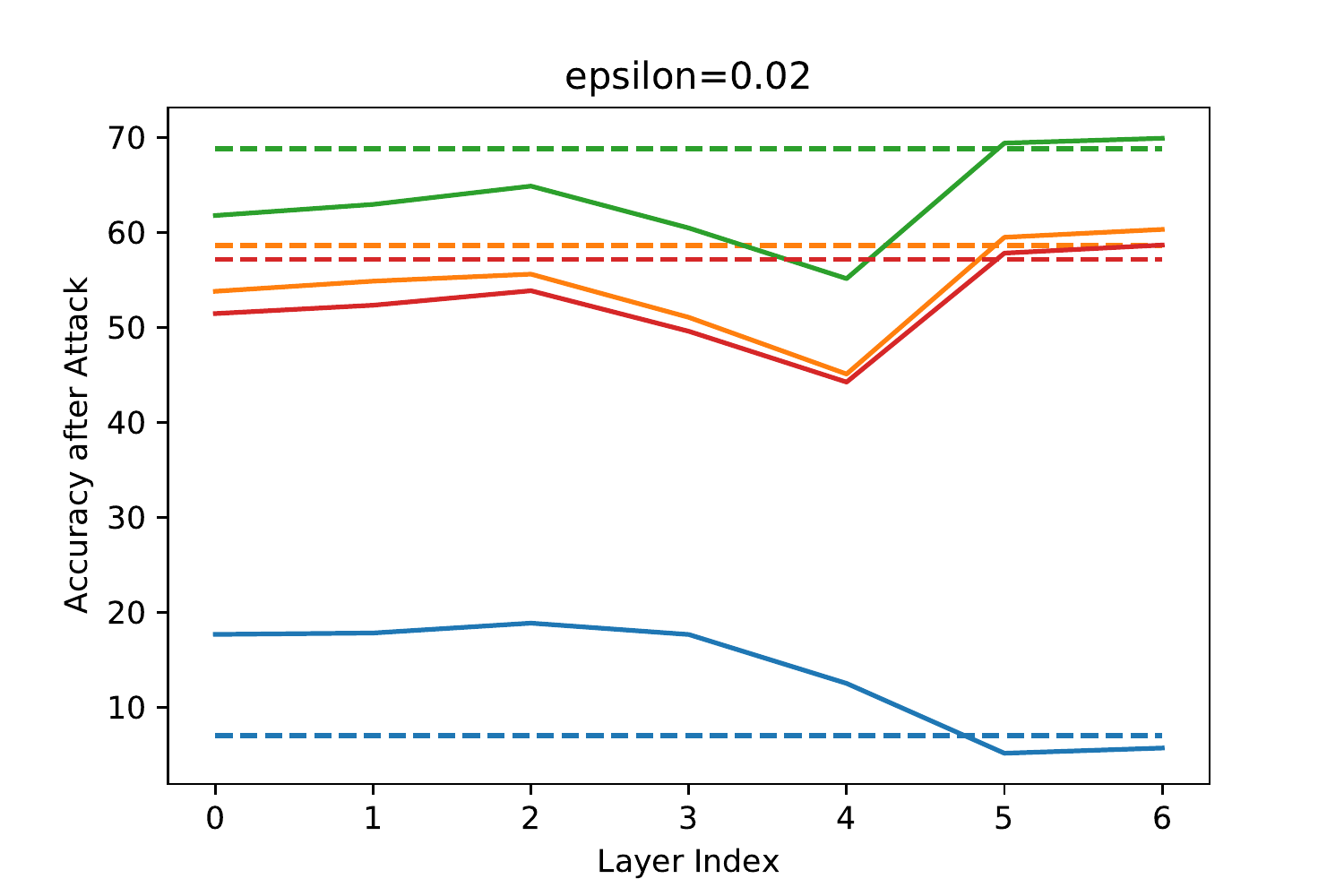}
  \end{minipage}
  \hspace{-0.1cm}
  \hfill
  \begin{minipage}[t]{0.45\textwidth}
    \centering
    \includegraphics[width=\textwidth]{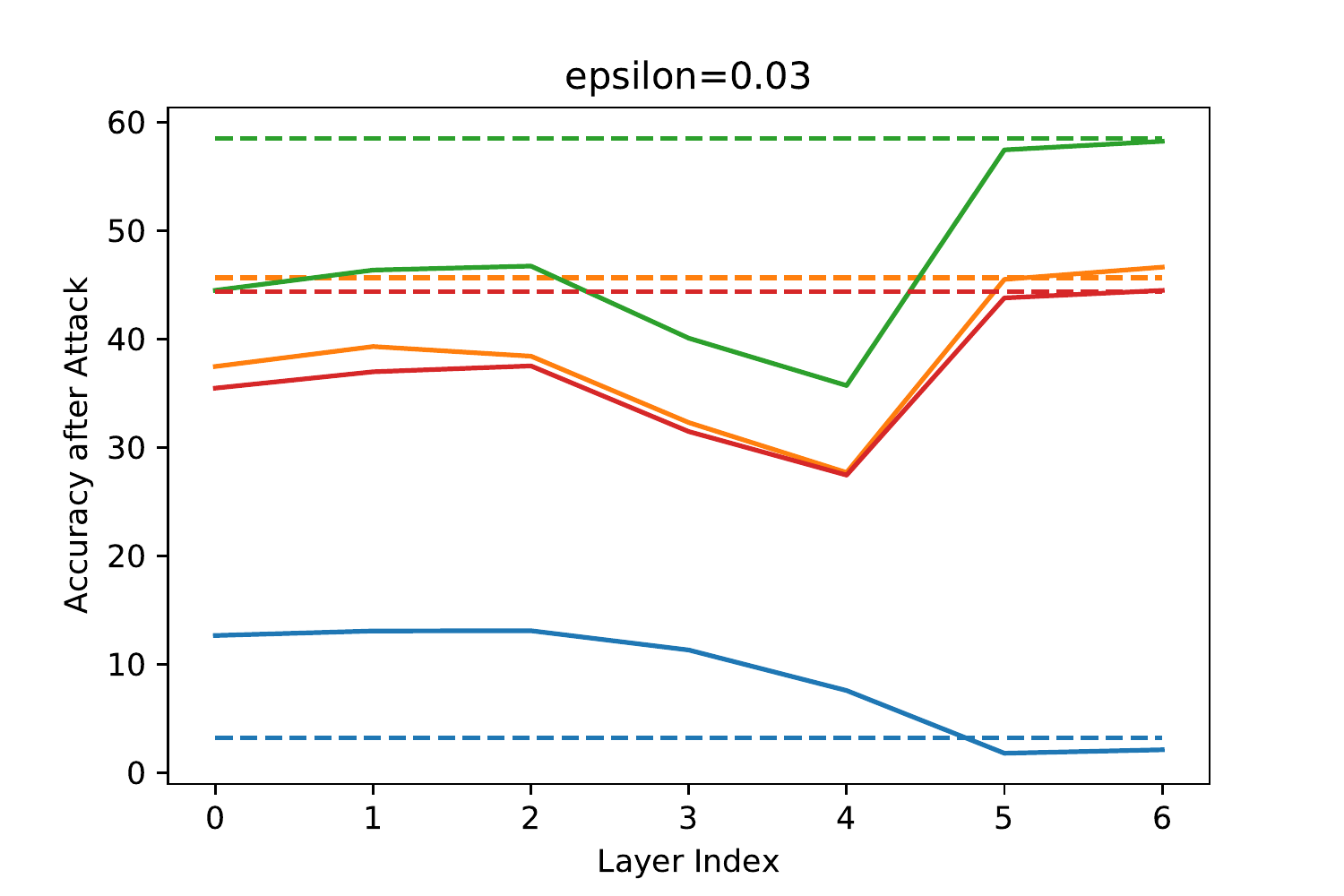}
  \end{minipage}
  \begin{minipage}[t]{0.45\textwidth}
    \centering
    \includegraphics[width=\textwidth]{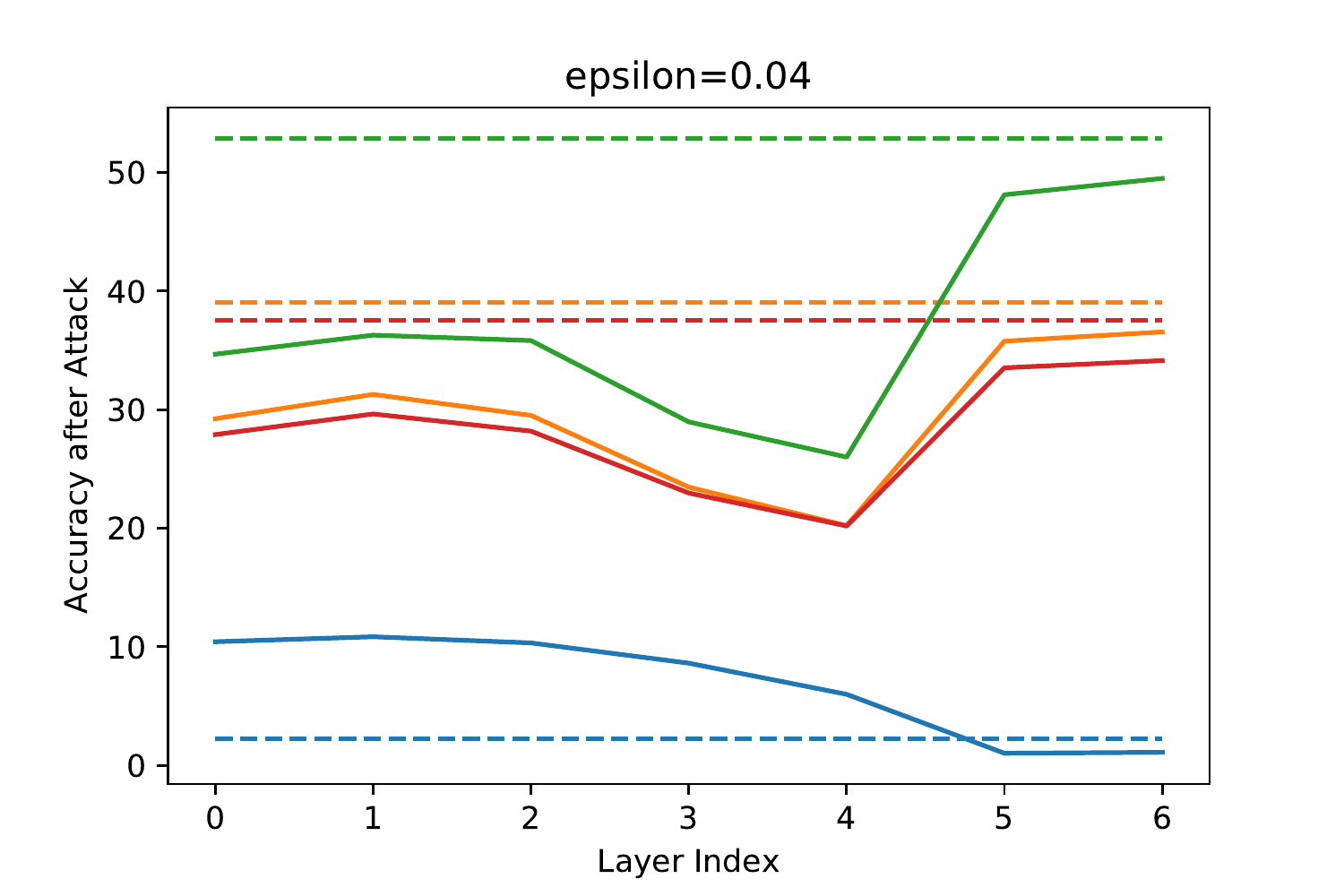}
  \end{minipage}
  \hspace{-0.1cm}
  \hfill
  \begin{minipage}[t]{0.45\textwidth}
    \centering
    \includegraphics[width=\textwidth]{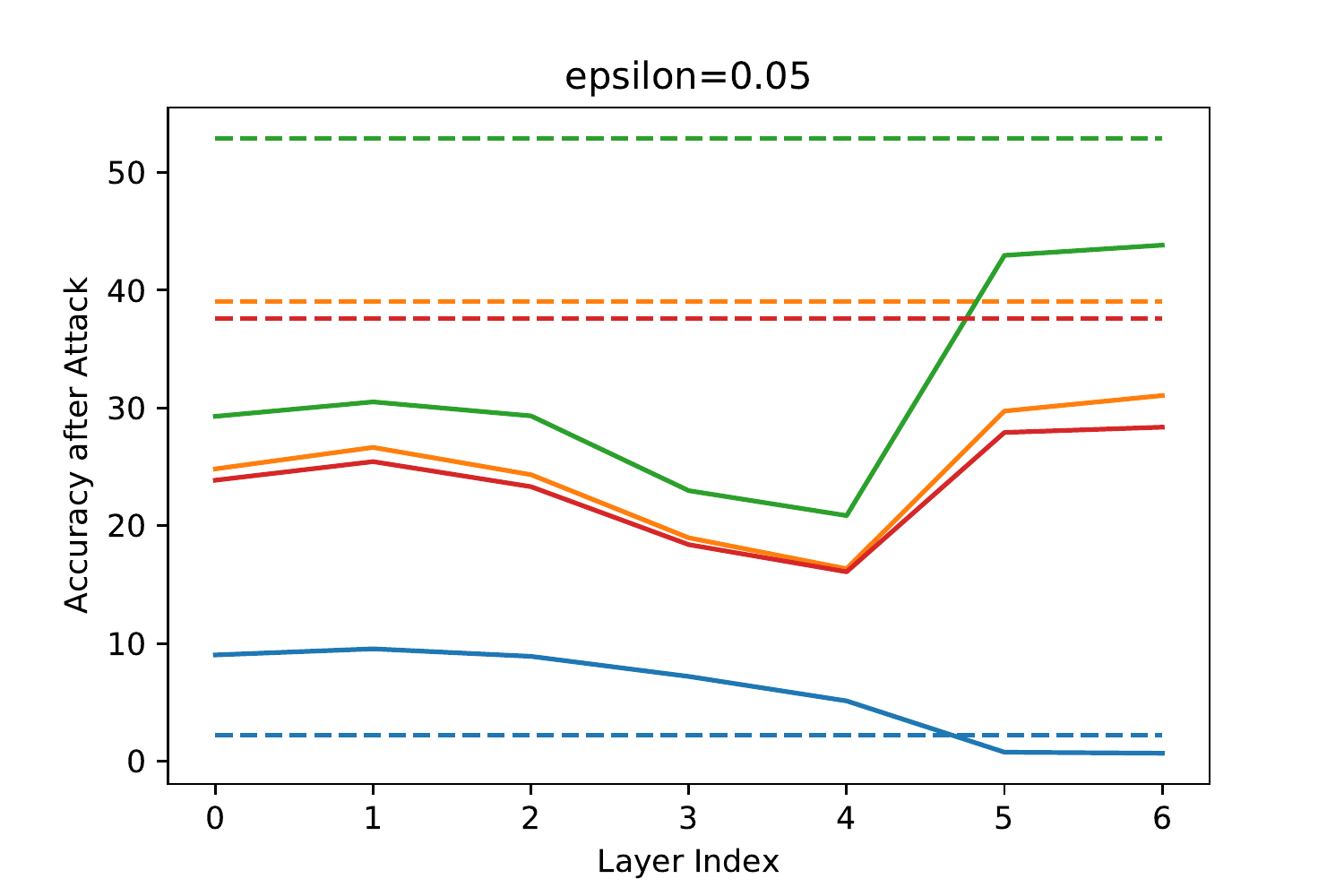}
  \end{minipage}
  \centering
  \includegraphics[width=0.2\textwidth]{appdx_figures/transferability_legend.pdf}
  \caption{\label{fig:epsilons} Transferability graphs for different epsilons}
\end{figure}

In this experiment, we use ILAP to generate adversarial examples with an I-FGSM baseline attack on ResNet18 with $\epsilon = 0.01, 0.015, 0.02, 0.025$, while other settings are kept the same as in section \ref{ILAP_cifar10}. We then evaluated their transferability against I-FGSM baseline on the adversarial examples of the whole test set, as shown in Figure \ref{fig:epsilons}.

\hspace{2em}

\section{Learning Rate Ablation}

We set iterations to 20 for both I-FGSM and I-FGSM with Momentum and experimented different learning rates on ResNet18. We then evaluate different models' accuracies on the generated $50 \times 32 = 1600$ adversarial examples, as shown in Table \ref{tab:ifgsmlr} and \ref{tab:momentumifgsmlr}.

\begin{table*}[!tbh]
  \caption{\label{tab:ifgsmlr} Learning rate ablation for I-FGSM }
  \centering
  \begin{tabular}{cccccc}
    \toprule
    learning rate &  ResNet18\footnote[2]{Model that is exactly the same model as the source model.} & SENet18 & DenseNet121 & GoogLeNet \\
    \midrule
    0.002  & 3.3\% & 44.9\% & 47.1\% & 59.3\% \\
    0.008  & 0.8\% & 45.6\% & 46.8\% & 60.0\% \\
    0.014  & 0.6\% & 47.2\% & 49.4\% & 59.5\% \\
    0.02 & 1.3\%  & 46.8\% & 51.4\% & 59.8\% \\
    \bottomrule
  \end{tabular}
  \end{table*}
  
  \begin{table*}[!tbh]
  \caption{\label{tab:momentumifgsmlr} Learning rate ablation for I-FGSM with Momentum}
  \centering
  \begin{tabular}{cccccc}
    \toprule
    learning rate &  $\text{ResNet18}^\dagger$ & SENet18 & DenseNet121 & GoogLeNet \\
    \midrule
    0.002  & 5.9\% & 35.0\% & 36.6\% & 46.1\% \\
    0.008  & 0.6\% & 43.0\% & 43.8\% & 56.1\% \\
    0.014  & 0.4\% & 43.6\% & 45.2\% & 55.9\% \\
    0.02 & 0.4\%  & 44.1\% & 46.4\% & 57.2\% \\
    \bottomrule
  \end{tabular}
  \end{table*}

\section{Comparison to TAP \cite{Zhou2018TransferableAP}}

CIFAR-10 \cite{Krizhevsky2009LearningML} results comparing a 20 iteration TAP \cite{Zhou2018TransferableAP} baseline to 10 iterations of ILAP using the output of a 10 iteration TAP attack are shown in Table \ref{tab:tap-result-cifar}.

\begin{table*}[!htb]
  \caption{\label{tab:tap-result-cifar} ILAP vs. TAP Results}
  \centering
  \begin{threeparttable}
\begin{tabular}{cccc}
\toprule
  & & \multicolumn{2}{c}{TAP \cite{Zhou2018TransferableAP}}  \\
    \cmidrule(r){3-4} 
    Source & Transfer & 20 Itr & Opt ILAP  \\
\midrule

  & $\text{ResNet18}^\dagger$  &                          6.2\% &           \textbf{1.9\%} (6)  \\
  ResNet18 & SENet18 &                                    31.6\% &          \textbf{28.4\%} (4)  \\
    ($l=4$) & DenseNet121 &                                32.7\% &          \textbf{28.5\%} (4)  \\
    & GoogLeNet &                                          41.6\% &          \textbf{36.8\%} (4)  \\
 
    \midrule
     &     ResNet18 &                                     31.4\% &           \textbf{23.5\%} (4) \\
    SENet18 &      $\text{SENet18}^\dagger$ &              2.0\% &            \textbf{1.7\%} (5) \\
     ($l=4$) &  DenseNet121 &                              31.3\% &          \textbf{24.1\%} (4)  \\
     &    GoogLeNet &                                      41.5\% &          \textbf{33.1\%} (4)  \\
    
    \midrule
  &     ResNet18 &                                         35.2\% &          \textbf{27.4\%} (6)  \\
 DenseNet121 &      SENet18 &                              34.2\% &          \textbf{26.8\%} (7)  \\
  ($l=6$)  &  $\text{DenseNet121}^\dagger$ &                4.8\% &           \textbf{1.0\%} (9)  \\
  &    GoogLeNet &                                         37.8\% &          \textbf{29.8\%} (6)  \\
 
  \midrule
     &     ResNet18 &                                    37.1\% &            \textbf{33.6\%} (9) \\
  GoogLeNet &      SENet18 &                              36.5\% &          \textbf{32.9\%} (9)  \\
    ($l=9$)  &  DenseNet121 &                              32.6\% &          \textbf{28.1\%} (9)  \\
  &    $\text{GoogLeNet}^\dagger$ &                        1.3\% &          \textbf{0.4\%} (12)  \\

\bottomrule
\end{tabular}
\begin{tablenotes}
  \item[$\dagger$] Same model as source model.
  \end{tablenotes}
\end{threeparttable}
\caption*{Table \ref{tab:tap-result-cifar}. Same as experiment in Table 2 of the main paper but with TAP. Hyperparameters for TAP are set to $lr =0.002, \epsilon = 0.015, \lambda = 0.005, \alpha = 0.5, s = 3, \eta = 0.01$.}
\end{table*}

\end{document}